# Vega-Fafnir: Is Greedy Decoding the Same Decoding? Where Rounding Noise in Speculative Verification Comes From and What Removing It Costs

*Qwen3-0.6B in bfloat16: cached, padded, batched, speculative and fixed-width paths against a width-1 reference, with a calibrated margin gate, four ways to resolve a flagged token, a width scan, a distribution shift and a float32 check*

Mohammed Nayeem Khan
mohammed.nayeem.khan@intel.com
mnknayeem@gmail.com

Navin Kumar R
navin.kumar.r@intel.com

Intel Corporation

**Abstract**

Serving engines treat the ways of running a greedy language model as interchangeable: cached or not, padded or not, alone or batched, one token or several verified at once (speculative decoding). In finite precision they round differently. We measure how often the greedy text of Qwen3-0.6B (bfloat16, 128 new tokens, GSM8K and ARC prompts) differs from a reference that decodes one token per pass. Verifying 4 guessed tokens per pass changes 60% of 150 prompts and padding changes 68%; every first divergence has a top-1 minus top-2 logit gap of 0.375 or less. A split of the noise shows that cache entries written by multi-token passes are about as noisy as the multi-token pass itself. A margin gate (fall back to one token when the gap is below a bound calibrated on separate GSM8K prompts) whose fallback reads that cache does not help: 58 of 100 prompts differ, against 61 of 100 for plain verification. Rebuilding the cache with one-token passes leaves 0 of 100 different but costs 7.2 model passes per token, against 1.0 for the reference. Fixed-width decoding makes drafted and plain decoding identical (0 of 100 prompts differ) at 0.75 passes per token. On shifted MMLU and HellaSwag prompts the rebuild gate leaves 0 of 60 different, plain verification 48 of 60. On 138 further and shifted prompts the bound was exceeded at 0.6% of positions and missed 0 of 309 flips. One GPU, one drafter, small models.

**Key findings (every number recomputed from the cached results)**

1. Reference rerun: 0.0% [0.0, 6.0] (0/60) of prompts differ, so the reference is deterministic on this machine.
2. 4-token verification: 60.0% [52.0, 67.5] (90/150) of 128-token outputs differ from the width-1 reference.
3. Padded to a multiple of 64: 68.0% [60.2, 74.9] (102/150) differ; batch of four: 69.3% [61.5, 76.2] (104/150) differ.
4. All 296 first divergences pooled over three paths have a reference margin of 0.375 or less; 10.9% of positions do.
5. Margin gate with a stale cache (eps = q99): 58.0% [48.2, 67.2] (58/100) differ, against 61/100 for plain verification.
6. Margin gate with the cache rebuilt by width-1 passes: 0.0% [0.0, 3.7] (0/100) differ, at 7.24 target passes per token.
7. Oracle trigger with a stale cache: 49.0% [39.4, 58.7] (49/100) differ, against 61/100 for plain verification.
8. Fixed-width 5 decoding, drafts against plain: 0 of 100 prompts differ; each differs from the width-1 reference on 66 and 66 of 100.
9. Rebuild gate with eps = 0.125: 0.0% [0.0, 3.7] (0/100) differ, against 61/100 for plain verification, at 3.55 target passes per token.
10. Rebuild gate with eps = 0 (exact ties only): 33.3% [21.4, 47.9] (15/45) differ, against 27/45 for plain verification, at 1.59 target passes per token.
11. Shifted MMLU and HellaSwag prompts: rebuild gate 0.0% [0.0, 6.0] (0/60) differ at 10.71 passes per token; plain verification 80.0% [68.2, 88.2] (48/60).
12. Shifted prompts, stale-cache gate: 68.3% [55.8, 78.7] (41/60) differ.
13. Width scan (20 prompts, 2459 positions): the most bit-identical positions between a wider pass and the width-1 pass is 0; widths 2 and 3 give identical logit rows, and widths 4, 5 and 6 give identical rows.
14. Noise bound (q99 = 0.375) on prompts that did not set it: S above the bound at 99 of 17178 positions; 0 of 309 flips had a verify margin above the bound.
15. Scope: one GPU and software stack, prompt-lookup drafts, Qwen3-0.6B (1.7B for a replication), 128 new tokens, 42 prompts that set the noise bound, at most 150 evaluation prompts per path.

## 1 Introduction

A greedy decoder is a function, so the same weights and the same prompt should always give the same tokens. Inference engines rely on that when they cache keys and values, pad sequences, batch requests, split long prompts into chunks or verify several guessed tokens in one pass. Each of these changes the shape of the matrix products on the GPU. In finite precision a different shape can add the same numbers in a different order and produce different low-order bits, and a tiny change in a logit can flip the argmax when the top two tokens are close. This paper asks how often mathematically identical decoding paths produce different greedy text, where the noise comes from, and what it costs to remove it.

The question matters for three groups. Evaluators who run a benchmark with greedy decoding assume that the score does not depend on the serving path. Reinforcement-learning pipelines assume that the sampler and the trainer assign the same token to the same context. Speculative decoding promises output identical to the target model (arXiv:2211.17192), and its drafters repeat the promise. The promise holds in exact arithmetic. On a GPU the verification pass scores k+1 positions at once while plain decoding scores one, so the two run different kernels.

The literature establishes that the effect exists (Section 2). What we could not find is a measurement of how often speculative verification at batch size 1 changes greedy text, a split of the noise by where it enters, or a test of whether a calibrated margin gate restores exactness. We study Qwen3-0.6B in bfloat16 against a reference that decodes one token per pass, with 128 new tokens per prompt, and check the main claims on Qwen3-1.7B where it ran.

**Findings.** (1) Verifying 4 prompt-lookup tokens per pass changes 60.0% [52.0, 67.5] (90/150) of prompts, padding 68.0% [60.2, 74.9] (102/150) and batching 69.3% [61.5, 76.2] (104/150); a rerun of the reference changes 0.0% [0.0, 6.0] (0/60). Every first divergence has a reference margin of 0.375 or less, where 10.9% of positions lie. (2) Cache entries written by wide passes are about as noisy as the wide pass itself, so a width-1 fallback over such a cache is not a reference computation. A margin gate whose fallback reads that cache leaves 58.0% [48.2, 67.2] (58/100) of prompts different, against 61/100 for plain verification on the same prompts. Rebuilding the cache with width-1 passes before the fallback leaves 0.0% [0.0, 3.7] (0/100), at 7.2 target passes per token against 1.0 for the reference. (3) Fixed-shape decoding, where every pass has the same width, makes drafted and plain decoding agree on 100 of 100 prompts at 0.75 passes per token, but not with the width-1 reference (66/100 differ). Held-out shifted prompts, bound coverage, a float32 comparison, cheap-path gates, a second model size and an answer-level check are in Sections 9 to 12.

**What is in-family.** The noise bound was set on 42 calibration prompts (all from GSM8K train). GSM8K evaluation prompts are in-family for that bound. ARC prompts, which are multiple-choice questions, and the shifted MMLU and HellaSwag prompts (Section 9) are out-of-family. The machine was shared with about eight other jobs, so runs are modest in size: 500 evaluation prompts are available and each path ran on between 60 and 150 of them (Table 1). Every claim below is limited to what that table lists.

## 2 Related work

Speculative decoding was introduced with the statement that outputs are identical to those of the target model (arXiv:2211.17192). Self-speculative decoding with layer skipping makes the same statement without an auxiliary model (arXiv:2309.08168), and SuffixDecoding drafts from suffix trees over earlier text (arXiv:2411.04975). These works concern the algorithm in exact arithmetic. No abstract we read reports how often an implementation's greedy output differs from plain decoding once rounding enters. The theory of acceptance in arXiv:2606.30265 also treats margins in exact arithmetic.

Work on reproducibility shows that batch size, GPU count and GPU type change bfloat16 greedy outputs, and that a reasoning model can drift by up to 9% in accuracy because early rounding differences cascade (arXiv:2506.09501). That paper computes in float32 with 16-bit weights. A comparison of five inference engines with fixed weights and hardware finds that the engine alone can shift benchmark scores by up to 16.6 percentage points (arXiv:2605.19537). In reinforcement learning, two engines with the same weights give different token probabilities, with argmax flips on Qwen3 in bfloat16, and the remedy tested is batch-invariant kernels with key-value splitting disabled (arXiv:2605.14220); float16 removes much of that mismatch (arXiv:2510.26788). A microbenchmark of Flash Attention finds about ten times the numeric deviation of baseline attention at bfloat16, growing with sequence length (arXiv:2405.02803). For batched speculative decoding, a forensic study traces corrupted outputs to desynchronised positions, masks and caches, reports 95% exact match to standard decoding at batch size 8 once those are fixed, and attributes the rest to floating-point non-determinism (arXiv:2510.22876). We look at that residual at batch size 1, where the bookkeeping is correct by construction, and split it by source.

LLM-42 decodes on a fast path and replays candidates under a fixed-shape reduction schedule in windows, committing the tokens that are consistent and rolling back the rest (arXiv:2601.17768). Its goal is run-to-run determinism under dynamic batching, and the parts of the paper we read do not report a logit margin or a split of the noise by source. Its cost figures for batch-invariant kernels (up to 63% lower GEMM throughput, up to 56% end-to-end overhead) are the reason to look for a cheaper certificate, which is what the margin gate here tries to be.

Margin-aware verification uses the target margin to accept runner-up tokens and gain speed (arXiv:2601.15498). Approximate speculative decoding accepts mismatches under a regret gate and a budget, and reduces to strict greedy verification at zero budget (arXiv:2608.03447). Both trade exactness for speed. The gate studied here uses the same signal in the other direction: it refuses a commitment when the margin is too small to trust. Early-exit decoding calibrates a confidence threshold on held-out data to bound the damage of leaving early (arXiv:2207.07061), and conformal language modeling calibrates rejection rules with a coverage guarantee (arXiv:2306.10193). Our calibration of a noise bound on held-out prompts follows the same recipe, and we test it on prompts from a different distribution.

What is new relative to this literature, as far as the abstracts and the sections we read show? We report four things. The first is a measurement of how often 4-token prompt-lookup verification, padding and batching change greedy text at batch size 1 against a width-1 reference. The second is a teacher-forced split of the rounding noise into the part from the query shape and the part from cache entries written by wide passes. The third is a test of whether a calibrated margin gate restores exactness, with several ways to resolve a flagged token and an oracle bound. The fourth is a fixed-shape decoder and a width scan as a check on the batch-invariance idea. We did not review engine source code, and a margin gate with cache rebuild may exist in engines or in papers that our searches (Hugging Face papers and arXiv) did not reach.

## 3 Setup: model, prompts, paths and noise statistic

Qwen3-0.6B (Apache-2.0) is loaded with transformers 5.20 and PyTorch 2.11 on one RTX 4090 that was shared with about eight other jobs. A second size, Qwen3-1.7B, is used for one replication. Prompts are rendered with the chat template with thinking switched off, and decoding is greedy with at most 128 new tokens (stop at an end-of-sequence token). TF32 is disabled so that float32 is true float32.

**Prompt sets.** The evaluation pool has 500 prompts. Of these, 250 come from GSM8K test (step-by-step arithmetic), 125 from ARC-Easy test and 125 from ARC-Challenge test (a question with options, answered with an explanation), in a fixed shuffled order. Every run uses the first n prompts of that order, so a smaller run is a subset of a larger one. The calibration list has 120 disjoint prompts, 60 from GSM8K train followed by 60 from ARC-Challenge test. The noise bound was set on the first 42 of them (all from GSM8K train), because the calibration run was stopped early, and the remaining 78 prompts of the list serve later as a held-out check of the bound. The GSM8K evaluation prompts are therefore *in-family* for the bound, while the ARC evaluation prompts differ from the prompts that set it in format and subject (a question with lettered options). The shifted set has 200 prompts from MMLU test (subject questions with four options) and HellaSwag validation (choose a continuation), a different distribution of topics and prompt format from all of these. Results on ARC and on the shifted prompts are the out-of-family tests of the bound.

**Reference path.** Batch size 1, a DynamicCache, one whole prefill pass, then width-1 decode steps with the default SDPA attention. All other paths are compared with it on the same prompts. It is one fixed point of comparison. We do not claim it is closer to an exact-arithmetic answer than any other path, and Section 10 measures how close each path is to a float32 model.

**Other paths.** Padded: the prompt is left-padded to a multiple of 64 with a mask and explicit positions. Batch of 4: four prompts are decoded together with left padding. k-token verification (k = 4 unless stated) uses prompt-lookup drafting: the proposed tokens are the continuation of the most recent earlier occurrence of the longest matching suffix n-gram (n up to 3). They are scored together with the last committed token in one pass of width k+1, the longest agreeing prefix is committed plus the verifier's own token after it, and the cache is cropped back to the committed length. Cache entries for the committed tokens are therefore written by the wide pass and kept.

**Noise statistic.** Take a path and the reference at the same position, both teacher-forced along the reference tokens, and let D be the reference logit minus the path logit. For the path's argmax a and each other token j in the path's top 16, the noise can move j above a by D_j - D_a. S is the largest of these values, clamped at zero. If the path margin (top-1 minus top-2) is at least S, no rival can overtake the argmax, and a flip needs S larger than the gap between a and the overtaking rival. A bound eps on S gives a gate: when the path margin is at most eps, the commit is not trusted and a fallback resolves it. A path that has not yet diverged reads the reference tokens, so teacher forcing along them is the situation at the first divergence. We set eps to the 99th percentile (q99), the 99.9th percentile (q999) or the maximum of S for 4-token verification on the calibration set; the values are q99 = 0.375; q999 = 0.625; max = 1.375 (n = 5253 positions from 42 prompts).

**Noise sources, teacher-forced.** Along the reference tokens we run five passes at every position. Spec is a wide pass over a cache written by wide passes, which is what speculative verification does. The q pass is a wide pass over the reference cache, so it carries query-shape noise only. The c pass is a width-1 pass over the wide-written cache, so it carries cache noise only, and it is what a gate fallback sees. The last two are one whole-sequence pass and a whole-sequence pass left-padded to a multiple of 64. A width-1 pass that rebuilds the reference cache reproduces the reference logits exactly (largest absolute difference 0.0 over all 42 prompts), so any difference of q or c from the reference comes from the shape or from the cache contents and not from a bug in the replay.

**Gate variants.** Plain verification commits every verified token. The gate falls back when the verify margin is at most eps. In *stale* mode the fallback is a width-1 pass over the cache as the wide passes left it. In *chunk* mode the current chunk is recomputed token by token first. In *rebuild* mode the generated part of the cache is recomputed with width-1 passes (all of it, or the last 16 entries in rebuild-16), so the fallback reads a cache equal to the reference cache. The *oracle* variants fall back exactly where the verify argmax differs from the reference, which no real gate can know, and give the best that any trigger could achieve in that mode. *Fixed-shape* decoding gives every pass after the prefill the same width W by filling unused rows with a filler token.

**Statistics.** A prompt diverges when its token sequence differs from the reference. The first divergence index is the first generated position where the tokens differ, and the margin at divergence is the reference's top-1 minus top-2 logit gap there, computed in float32 from the stored logits. Rates carry 95% Wilson intervals over prompts, means carry bootstrap intervals over prompts, and comparisons between paths on the same prompts use the exact McNemar test. Greedy decoding uses no random seeds, so the only sampling is the choice of prompts. Tables use the prompts finished by every path they compare.

**Table 1.** Everything that ran. n is the number of prompts. Prompts are the first n of one shuffled order for each set, so a smaller run is a subset of a larger one.

| run | model | prompt set | n | new tokens | purpose |
|---|---|---|---|---|---|
| reference (width 1) | 0.6B | evaluation | 150 | 128 | comparison point for every path |
| reference rerun | 0.6B | evaluation | 60 | 128 | determinism control |
| 4-token verification | 0.6B | evaluation | 150 | 128 | speculative path |
| padded to 64 | 0.6B | evaluation | 150 | 128 | padding path |
| batch of 4 | 0.6B | evaluation | 150 | 128 | batching path |
| teacher-forced noise sources | 0.6B | calibration | 42 | 128 | five noise sources, noise bound |
| float32 teacher-forced | 0.6B | calibration | 43 | 128 | closeness to float32 |
| gate stale q99 | 0.6B | evaluation | 100 | 128 | fallback variant |
| gate chunk q99 | 0.6B | evaluation | 60 | 128 | fallback variant |
| gate rebuild-16 q99 | 0.6B | evaluation | 60 | 128 | fallback variant |
| gate rebuild q99 | 0.6B | evaluation | 100 | 128 | fallback variant |
| gate rebuild q999 | 0.6B | evaluation | 45 | 128 | fallback variant |
| gate rebuild eps 0.25 | 0.6B | evaluation | 45 | 128 | fallback variant |
| gate rebuild eps 0.125 | 0.6B | evaluation | 100 | 128 | fallback variant |

| run | model | prompt set | n | new tokens | purpose |
|---|---|---|---|---|---|
| gate rebuild eps 0 | 0.6B | evaluation | 45 | 128 | fallback variant |
| oracle stale | 0.6B | evaluation | 100 | 128 | fallback variant |
| oracle rebuild | 0.6B | evaluation | 30 | 128 | fallback variant |
| fixed width 5, drafts | 0.6B | evaluation | 100 | 128 | fixed-shape decoding |
| fixed width 5, plain | 0.6B | evaluation | 100 | 128 | fixed-shape decoding |
| teacher-forced noise sources | 0.6B | further calibration-set prompts | 78 | 128 | held-out check of the noise bound |
| teacher-forced noise sources | 0.6B | shifted | 60 | 128 | noise bound on shifted prompts |
| width scan (bit equality) | 0.6B | evaluation | 20 | 128 | which widths share kernels |
| reference | 0.6B | shifted | 60 | 128 | held-out shift |
| 4-token verification | 0.6B | shifted | 60 | 128 | held-out shift |
| gate stale q99 | 0.6B | shifted | 60 | 128 | held-out shift |
| gate rebuild q99 | 0.6B | shifted | 60 | 128 | held-out shift |
| early exit and 3 to 8 bit weights | 0.6B | evaluation | 100 | 128 (teacher-forced) | cheap paths |
| reference | 1.7B | evaluation | 45 | 128 | second model |
| 4-token verification | 1.7B | evaluation | 45 | 128 | second model |
| padded to 64 | 1.7B | evaluation | 45 | 128 | second model |
| wall-clock timing of seven paths | 0.6B | evaluation | 12 | 128 | time per token, context only |
| teacher-forced noise sources | 1.7B | calibration | 40 | 128 | noise split on the second model |
| gate stale, eps 0.375 | 1.7B | evaluation | 45 | 128 | gates on the second model |
| gate rebuild, eps 0.375 | 1.7B | evaluation | 45 | 128 | gates on the second model |

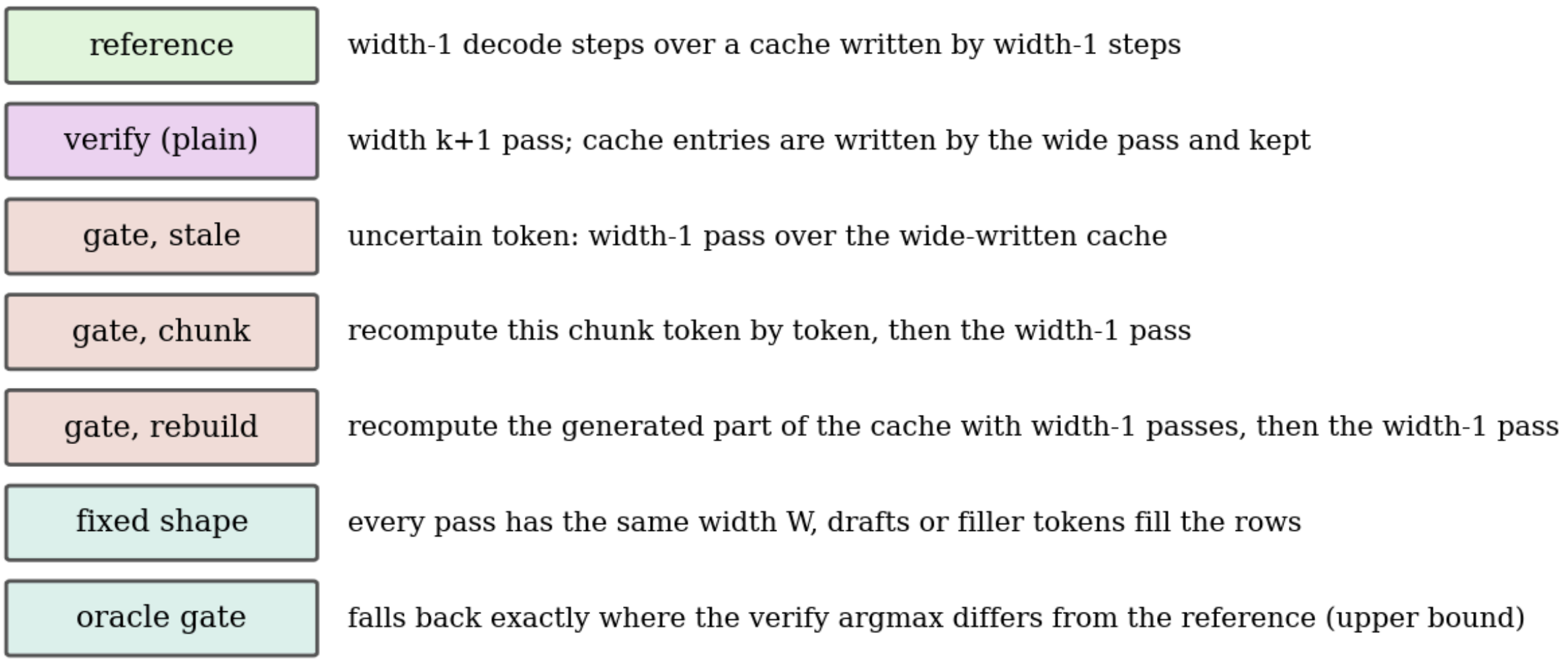

**Figure 1.** The execution paths compared. All compute the same greedy function. They differ in the shape of the matrix products (query rows per pass) and in which pass wrote each cache entry.

## 4 How often does greedy output differ?

Table 2 and Figure 2 give the headline of Part 1. The reference was run on 150 evaluation prompts (mean length 125 generated tokens, 124 of them cut at the 128-token limit); the other paths ran on the first prompts of the same order. After 128 new tokens, 4-token verification changed the output of 60.0% [52.0, 67.5] (90/150) of the 150 prompts. Padding to a multiple of 64 changed 68.0% [60.2, 74.9] (102/150). A batch of four left-padded prompts changed 69.3% [61.5, 76.2] (104/150). A second run of the reference itself changed 0.0% [0.0, 6.0] (0/60), so the reference is deterministic on this machine and every difference below comes from a different execution path. The share that had diverged by position 64 was 32.0% for 4-token verification, against 60.0% by position 128. Figure 3 shows the share still identical falling with position, and Figure 4 shows the hazard of a first divergence per token still at risk. The rate therefore depends on the generation length, and no single figure describes a path.

**Table 2.** Divergence from the width-1 reference in bfloat16, 128 new tokens. The last column counts prompts where only this path differs or only 4-token verification differs on the shared prompts, with the exact McNemar p. The last two numeric columns are medians over prompts that diverged.

| path | prompts | differ | share that differ [95% Wilson] | median first position | median reference margin | only this path / only 4-token verify |
|---|---|---|---|---|---|---|
| reference rerun | 60 | 0 | 0.0% [0.0, 6.0] | - | - | - |
| 4-token verification | 150 | 90 | 60.0% [52.0, 67.5] | 62 | 0.125 | - |
| padded to 64 | 150 | 102 | 68.0% [60.2, 74.9] | 34 | 0.125 | 27 / 15, p = 0.088 |
| batch of 4 | 150 | 104 | 69.3% [61.5, 76.2] | 44 | 0.125 | 27 / 13, p = 0.038 |

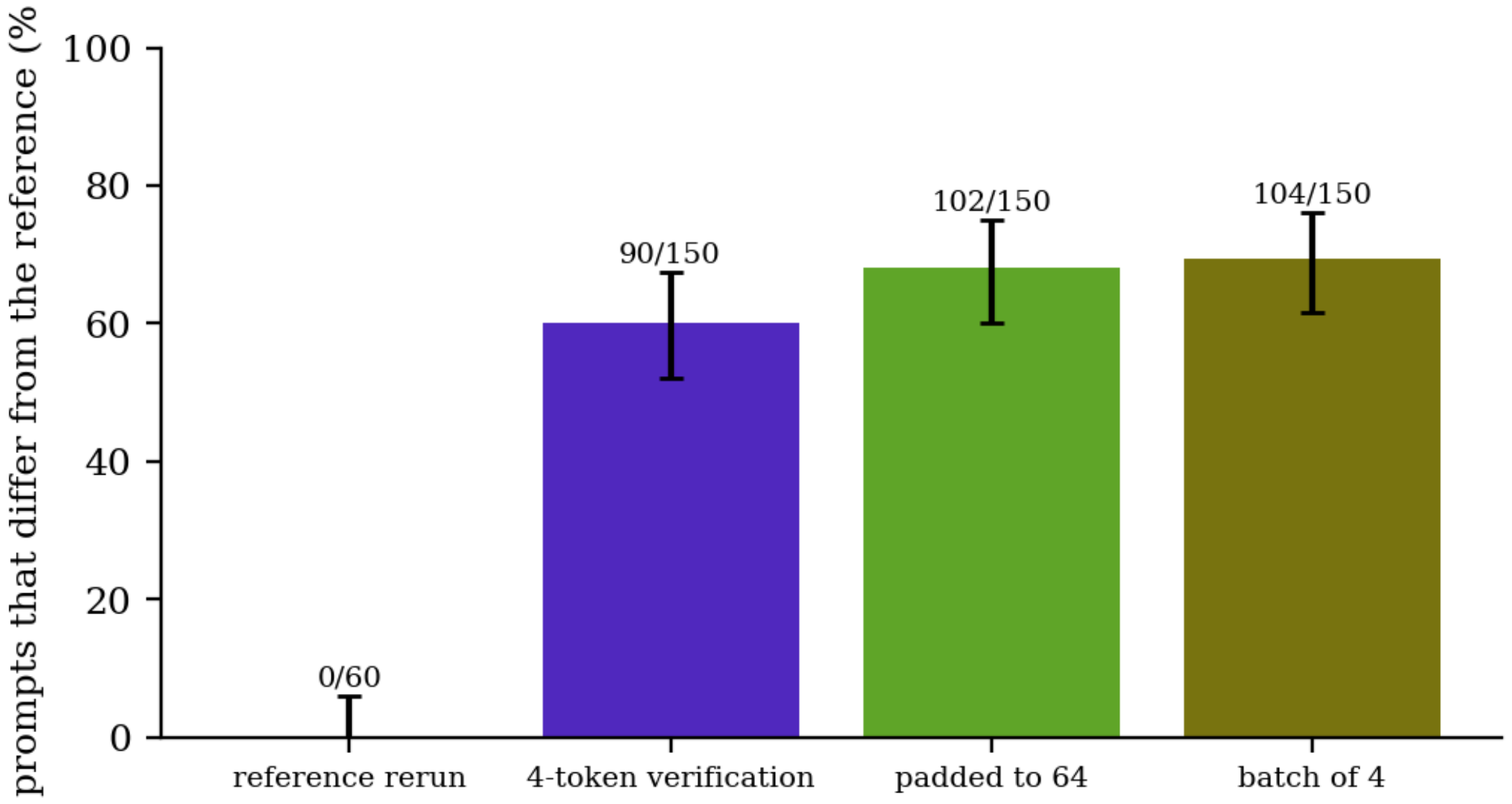

**Figure 2.** Share of evaluation prompts (128 new tokens) whose output differs from the width-1 reference, with 95% Wilson intervals; the count and number of prompts are above each bar. The rerun of the reference itself is the determinism control.

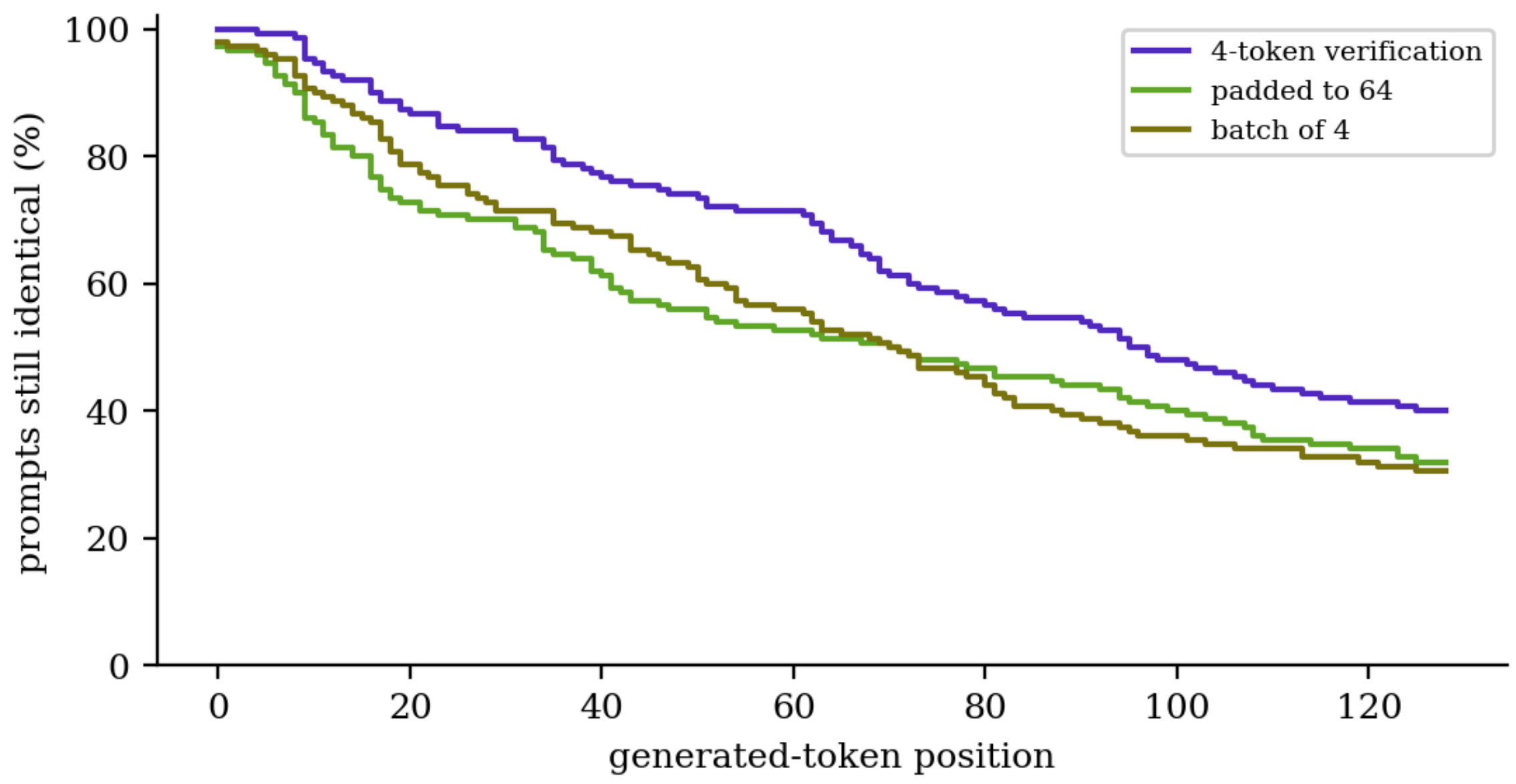

**Figure 3.** Share of prompts whose output is still identical to the reference after each generated position. The curves fall steadily, with no knee at a particular position.

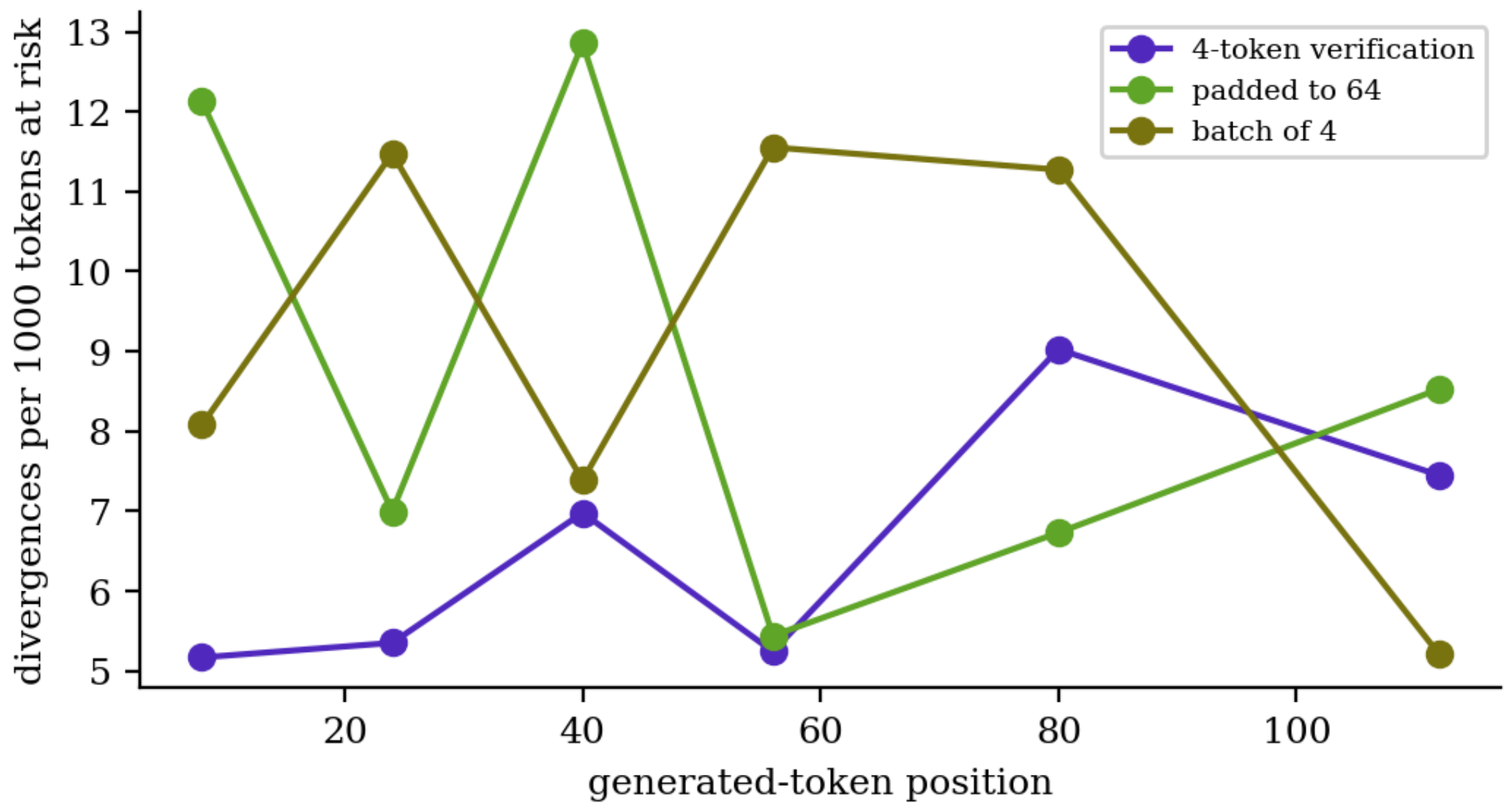

**Figure 4.** Hazard of first divergence by position: events per 1000 prompt-tokens still at risk.

The rate depends on the prompt source (Figure 5). For 4-token verification, 41.1% [30.5, 52.6] (30/73) of GSM8K prompts differ, against 81.4% [67.4, 90.3] (35/43) for ARC-Easy and 73.5% [56.9, 85.4] (25/34) for ARC-Challenge. The reference margin is at most 0.25 at 5.6% of GSM8K positions, 10.2% of ARC-Easy positions and 10.8% of ARC-Challenge positions, which is the density of positions where a one-step change in a bfloat16 logit can flip the argmax. The order of the divergence rates follows the order of that density.

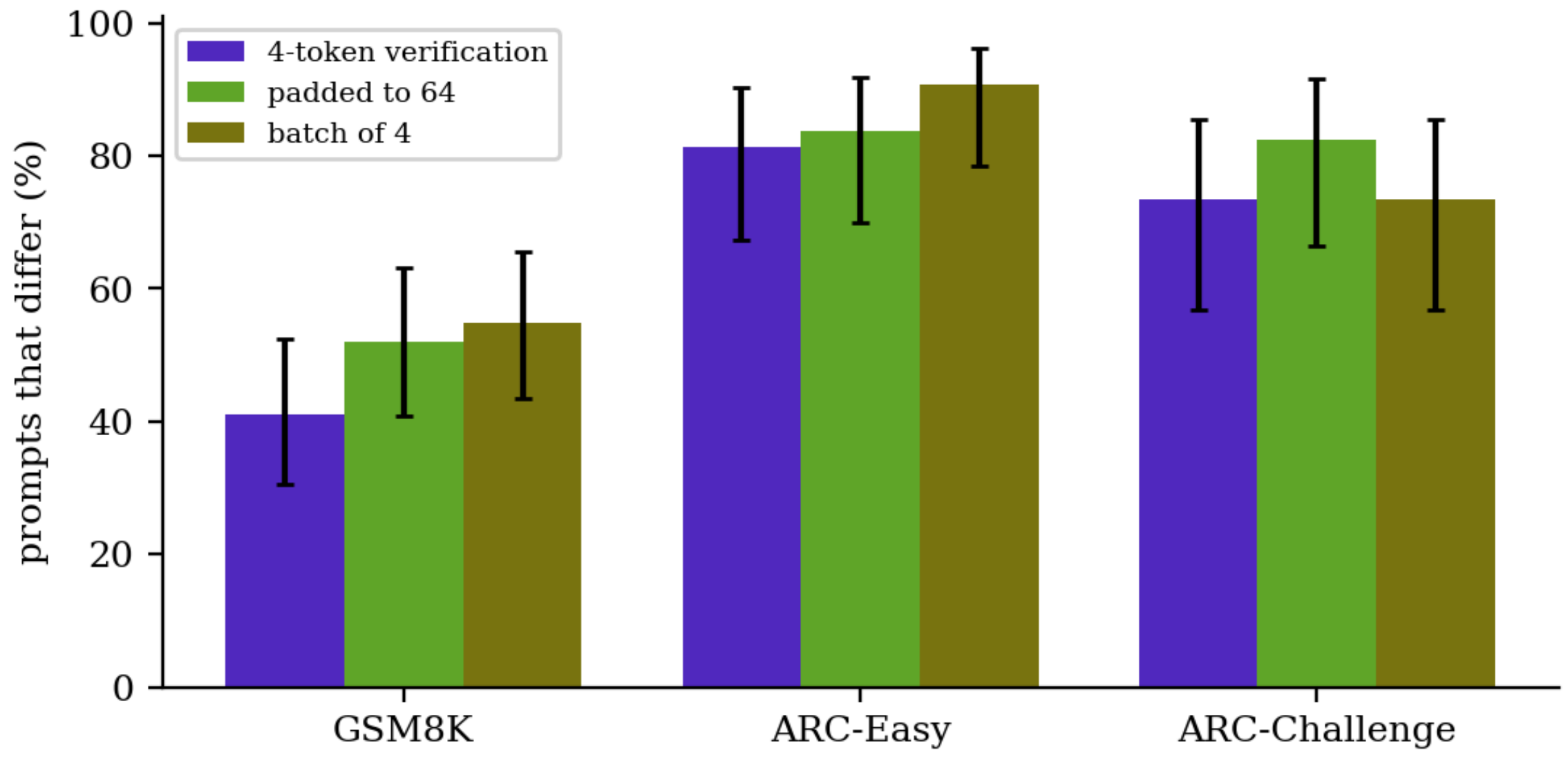

**Figure 5.** Share of prompts that differ from the reference, by prompt source, with 95% Wilson intervals.

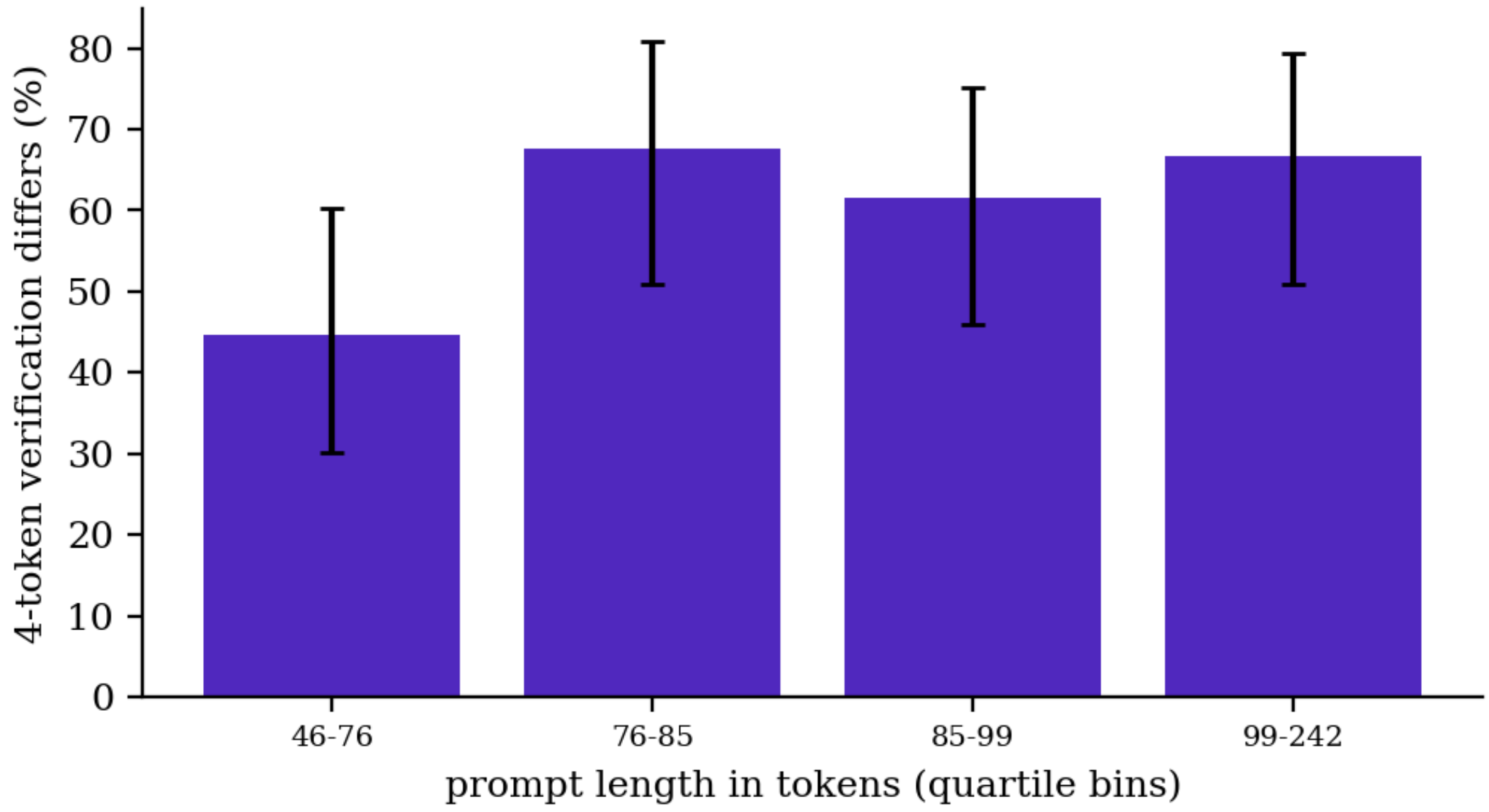

**Figure 6.** Divergence rate of 4-token verification by prompt length quartile, with 95% Wilson intervals.

## 5 Where it happens: margin and position

A divergence needs a near tie: if the reference has a wide gap between its top two tokens, a rounding difference of a few logit steps cannot reverse the order. Figure 7 compares the reference margin at the first divergence with the margin at all 18689 reference positions. Pooled over 3 paths there are 296 first divergences from 127 distinct prompts, with a median reference margin of 0.125, a maximum of 0.375 and 99.7% at 0.25 or less. In the same units 8.1% of all positions have a margin of 0.25 or less and the median margin over all positions is 3.12. The share of divergences at an exact tie in the stored bfloat16 logits (margin 0) is 39.2%, against 1.5% of all positions. The smallest non-zero margin on these prompts is 0.1250, and the next values are 0.1875, 0.2500, 0.3750. This is the grid of bfloat16, which keeps 8 significant bits, so logits between 16 and 32 lie 0.125 apart. A one-step change of either of the top two logits can reverse a gap of one step or close an exact tie. The 0.25 threshold was read off the 4-token verification divergences, whose largest reference margin is 0.250; 1 of the 296 pooled divergences lies above it, so it describes these data and is not an independent test.

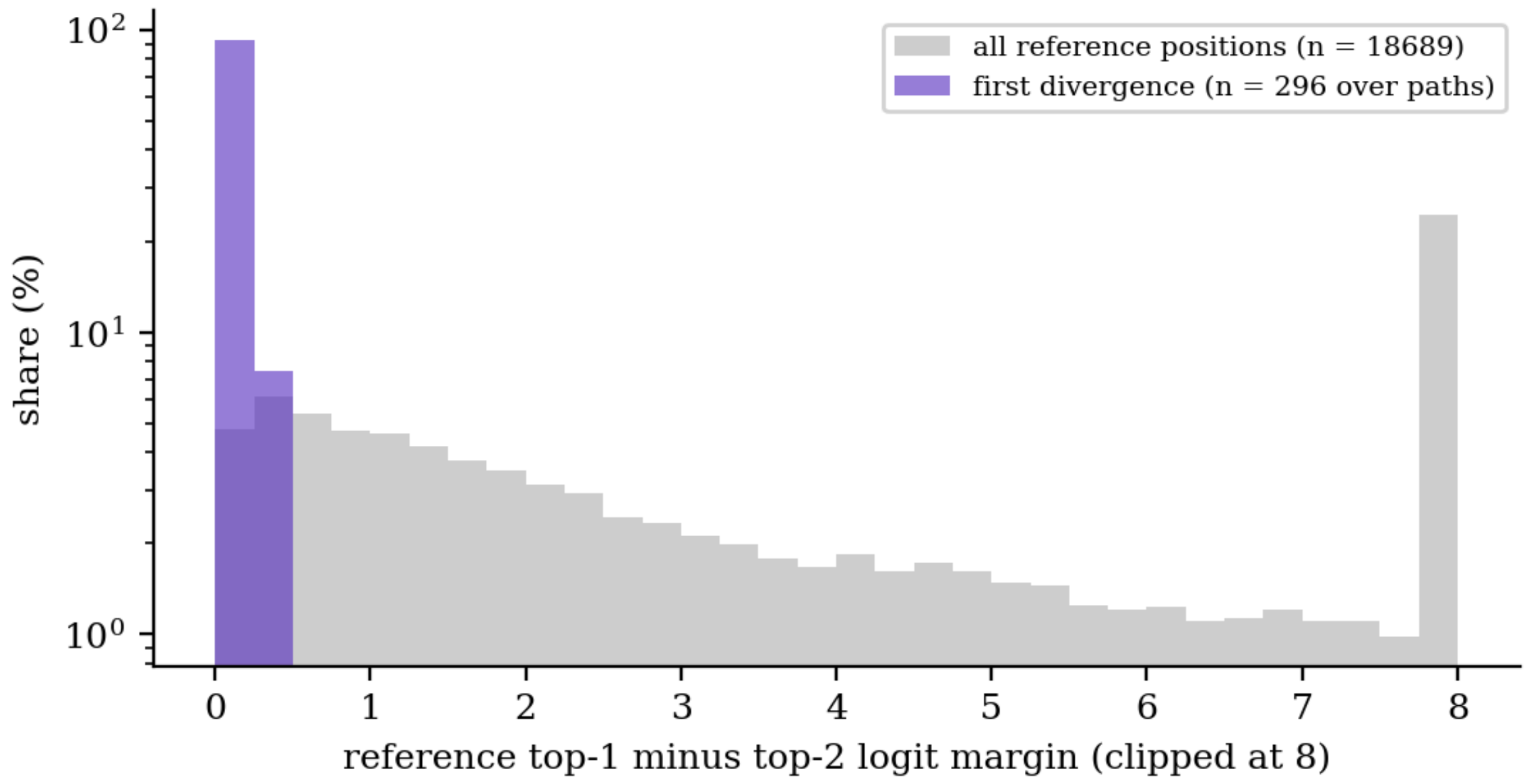

**Figure 7.** Reference margin at the first divergence compared with the margin at all positions (log scale). Divergences concentrate at the lowest margins; the largest margin at a divergence is 0.375.

Which signal ranks the fragile positions best? On the 42 teacher-forced calibration prompts we label a position positive when any of five noise sources (Section 6) flips the argmax there, and score it by the reference margin, the wide-pass margin, the reference entropy or the position (Figure 8). The area under the ROC curve is 0.993 [0.989, 0.996] for reference margin (low), 0.992 [0.989, 0.995] for wide-pass margin (low), 0.922 [0.910, 0.932] for reference entropy, 0.446 [0.388, 0.522] for position (bootstrap intervals over prompts). The best of the four is the reference margin (low). Figure 9 shows the price of a margin threshold: the share of positions that it sends to a fallback against the share of flips it catches.

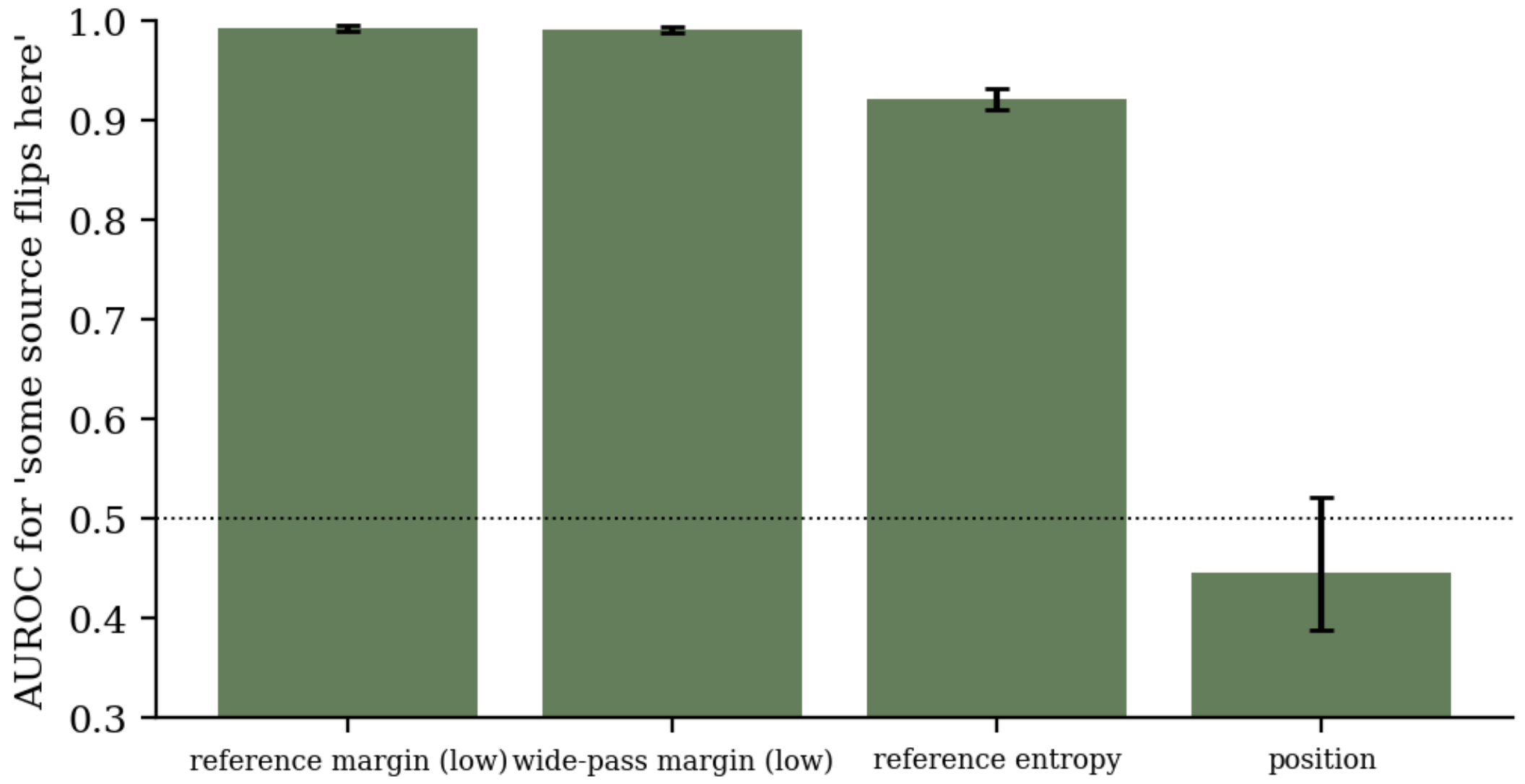

**Figure 8.** How well each signal ranks the positions where any of the five noise sources flips the argmax (71 of 5253 positions). Intervals are bootstrap over prompts.

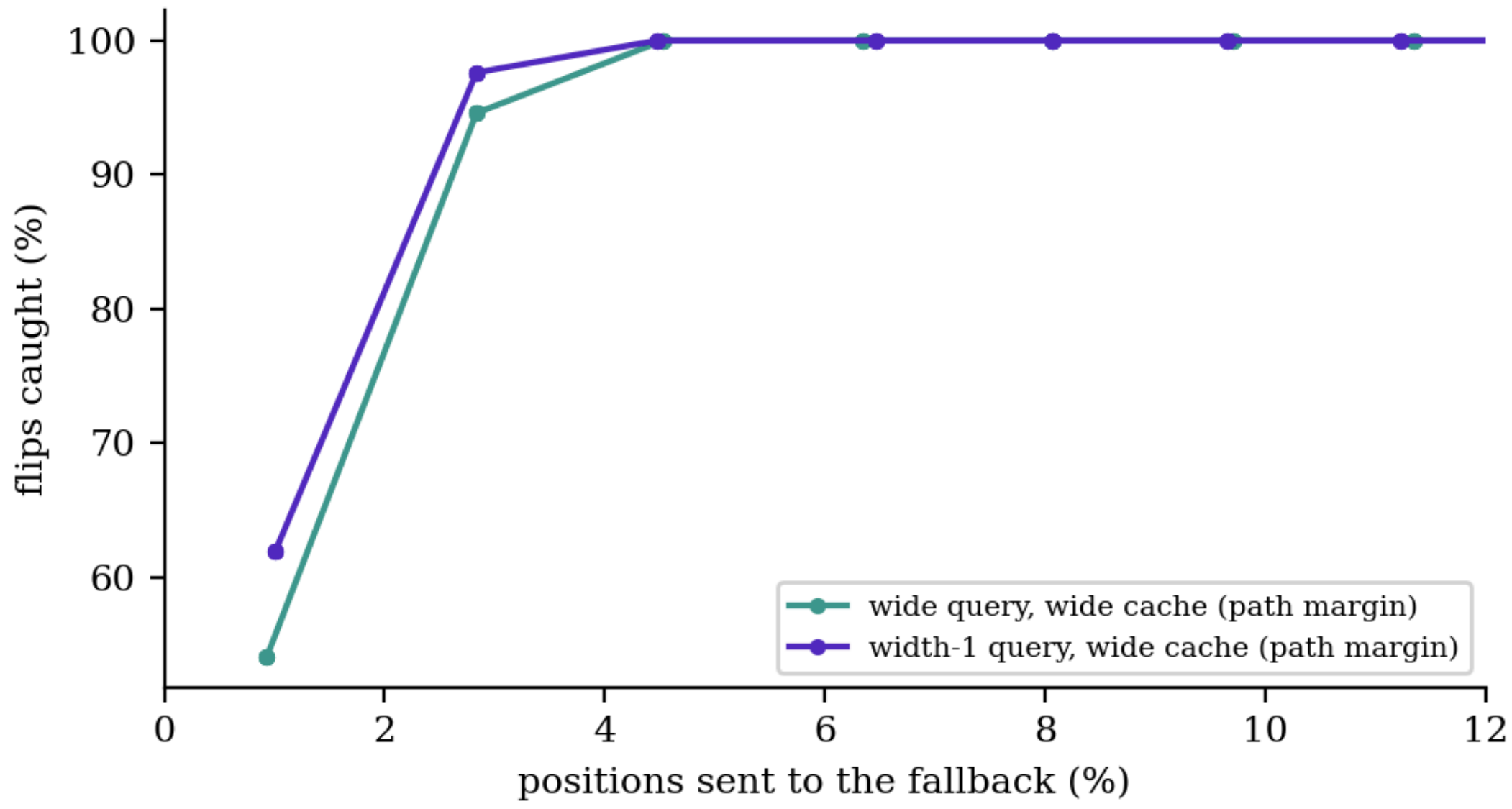

**Figure 9.** Operating curve of a margin threshold: share of positions flagged against share of argmax flips whose path margin lies at or below the threshold (thresholds 0 to 1.5).

A second, calibration-free signal is disagreement between paths. At 53 of the 5253 calibration positions at least two of the five noise sources pick different argmaxes, and at 94.3% of them the reference margin is 0.25 or less, against 4.5% of all positions. Disagreement is precise but incomplete: it covers 53 of the 71 positions where some source differs from the reference, because the reference can differ from all the other sources at once.

## 6 Where the noise comes from: query shape or cache

Why does a width-1 fallback not restore the reference? A speculative step differs from the reference in two ways. Its pass has more query rows, which changes the matrix-product shapes, and the cache entries it writes for the committed tokens come from that wide pass and not from width-1 passes. Table 3 separates the two along the reference tokens of 42 calibration prompts (5253 positions). A wide pass over the reference cache isolates the query shape (q). A width-1 pass over the wide-written cache isolates the cache contents (c). The check is exact: a width-1 pass that replays the reference cache reproduces the reference logits with a largest absolute difference of 0.0 over all prompts, so a non-zero S comes from the shape or the cache and not from the replay.

**Table 3.** Noise of five sources against the reference, teacher-forced along the reference tokens of the calibration prompts. Flips are argmax differences (per 1000 positions in brackets). S is the largest amount by which the noise can move a rival token above the path argmax; the mean is over per-prompt means with a bootstrap interval over prompts.

| source | positions | argmax flips | S > 0 | mean S [95% bootstrap] | S q99 | S q999 | S max |
|---|---|---|---|---|---|---|---|
| wide query, reference cache | 5253 | 40 (7.6) | 53.0% | 0.078 [0.076, 0.080] | 0.375 | 0.500 | 1.062 |
| width-1 query, wide cache | 5253 | 42 (8.0) | 72.7% | 0.117 [0.113, 0.120] | 0.375 | 0.625 | 0.938 |
| wide query, wide cache | 5253 | 37 (7.0) | 73.7% | 0.120 [0.116, 0.124] | 0.375 | 0.625 | 1.375 |
| whole sequence | 5253 | 43 (8.2) | 78.6% | 0.145 [0.139, 0.150] | 0.500 | 0.906 | 1.750 |
| whole sequence, padded | 5253 | 46 (8.8) | 78.0% | 0.143 [0.136, 0.149] | 0.500 | 0.625 | 1.250 |

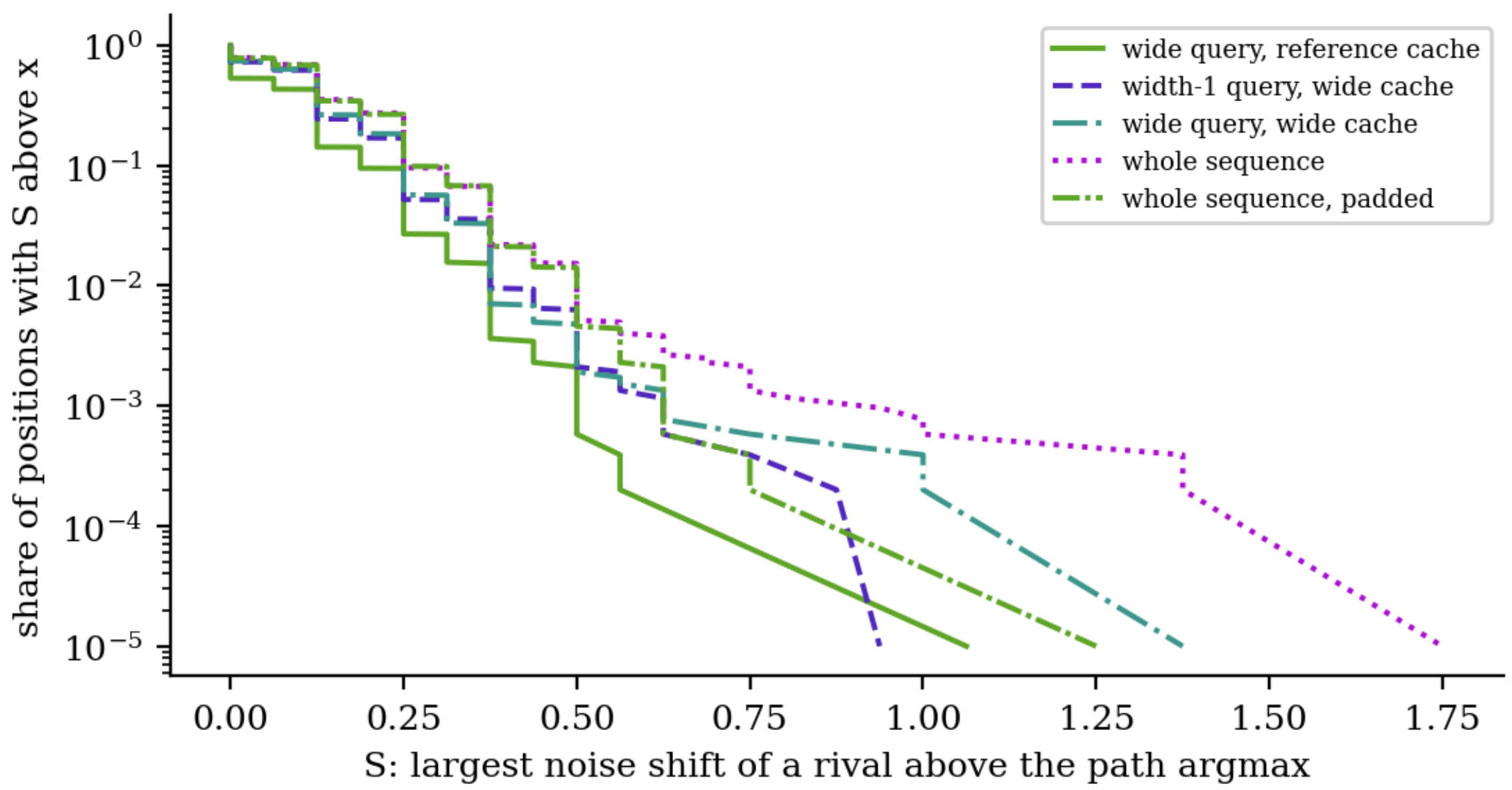

**Figure 10.** Upper tail of the noise statistic S for five noise sources, teacher-forced along the reference tokens of 42 calibration prompts (5253 positions).

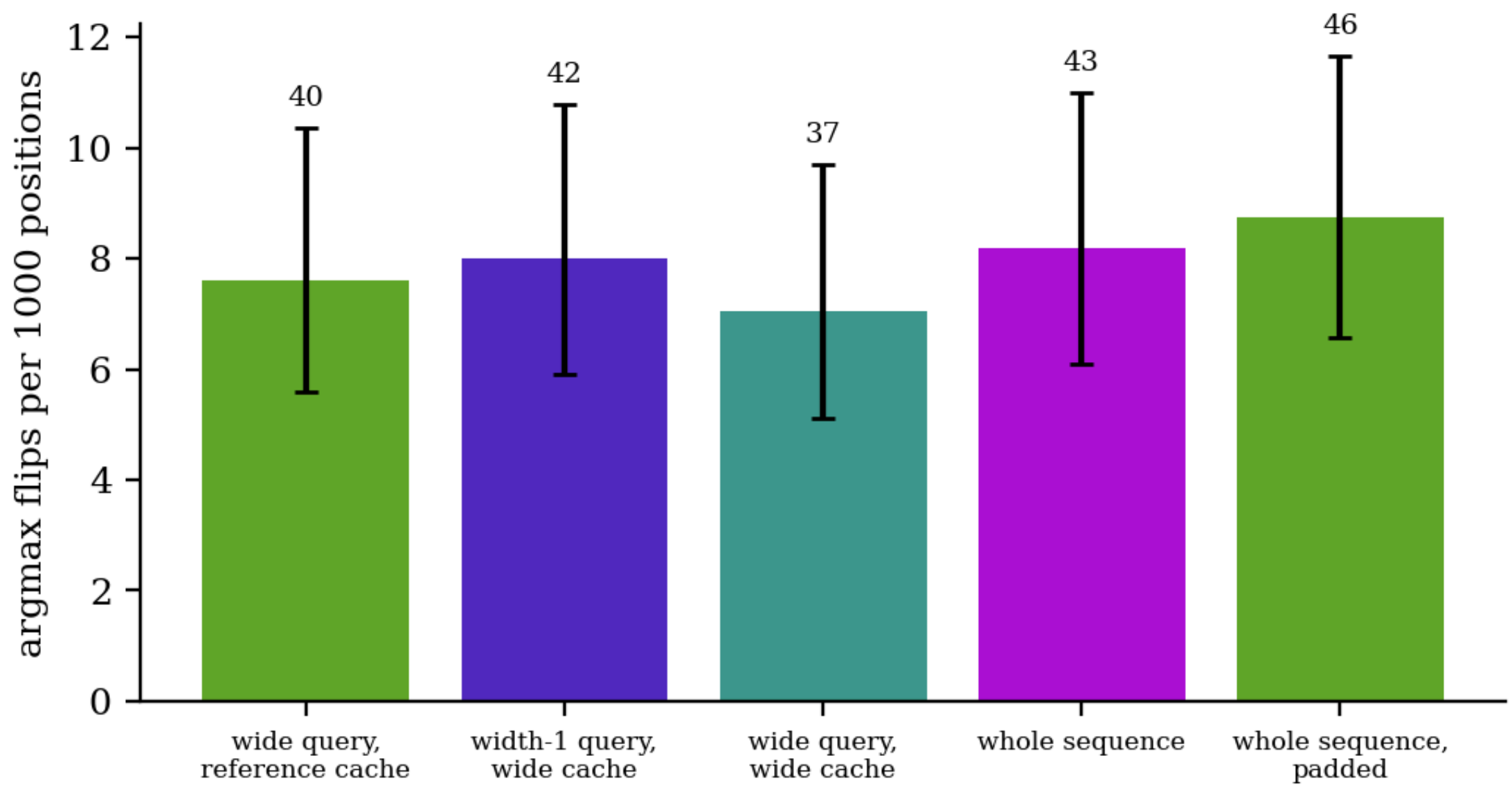

**Figure 11.** Argmax flips against the reference at the same position, per noise source (counts above the bars, 95% Wilson intervals over positions).

The cache source is not smaller than the query source. The mean of S is 0.117 for c, 0.078 for q and 0.120 for both together; the paired difference of per-prompt means between c and q is 0.039 [0.036, 0.042] (bootstrap over prompts). In argmax flips the sources are close: 40 for q, 42 for c and 37 for both together among 5253 positions, against 43 for a whole-sequence pass and 46 for the padded whole-sequence pass. Figure 12 shows the overlap: of the 37 positions where the verify pass flips, 29 also flip under q, 33 also flip under c and 27 under both, and 19 positions flip under q or c but not under the verify pass. The sources do not add up, which fits a flip being decided by whether a near tie is crossed and not by the size of the perturbation. What matters for a gate is that a width-1 fallback that reads a wide-written cache (c) flips the argmax about as often as the wide pass it is meant to correct. Figure 13 shows the mean of S by generated position for q and c.

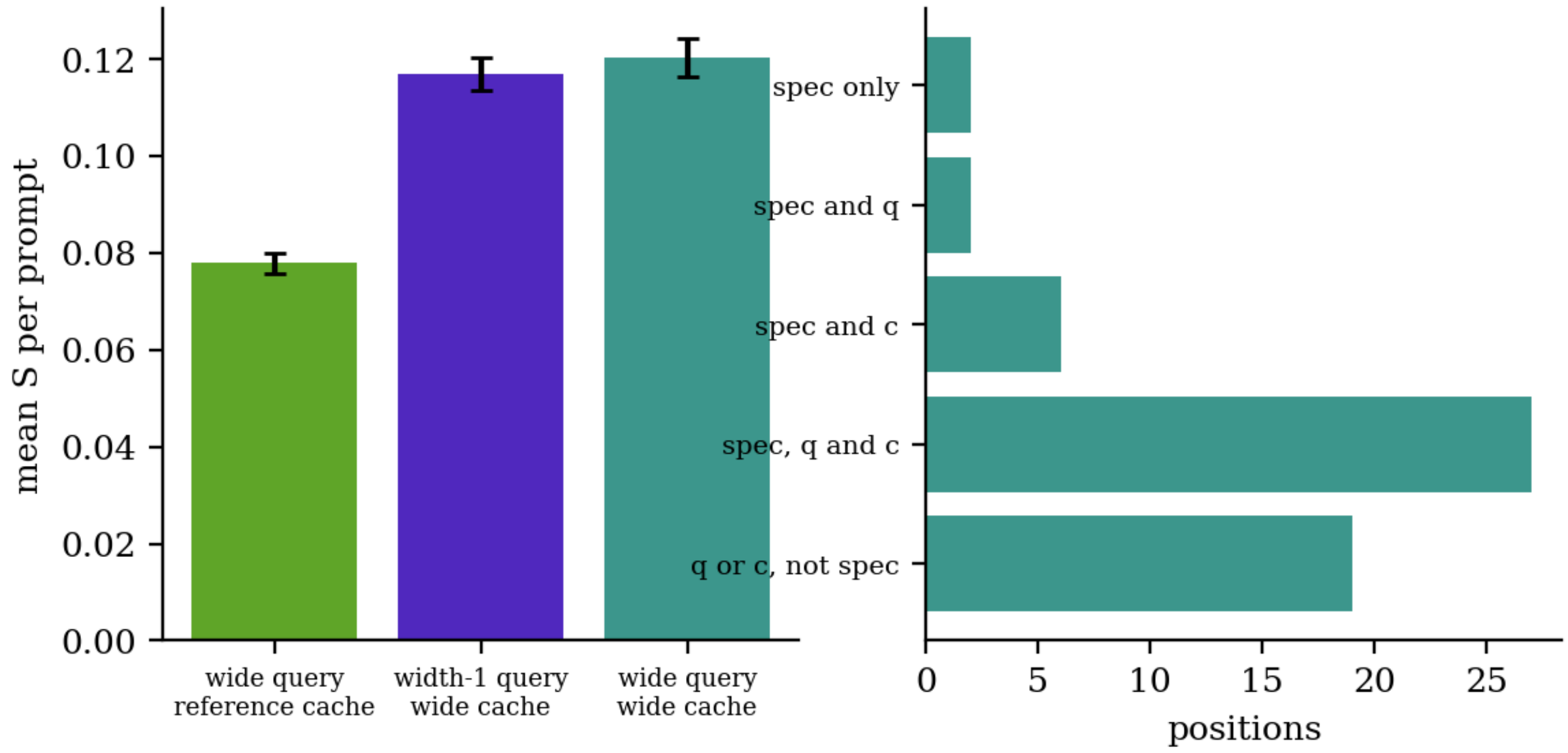

**Figure 12.** Left: mean of S per prompt for the query-shape source, the cache source and both together, with bootstrap intervals over prompts. Right: positions where the argmax flips, by which sources flip together.

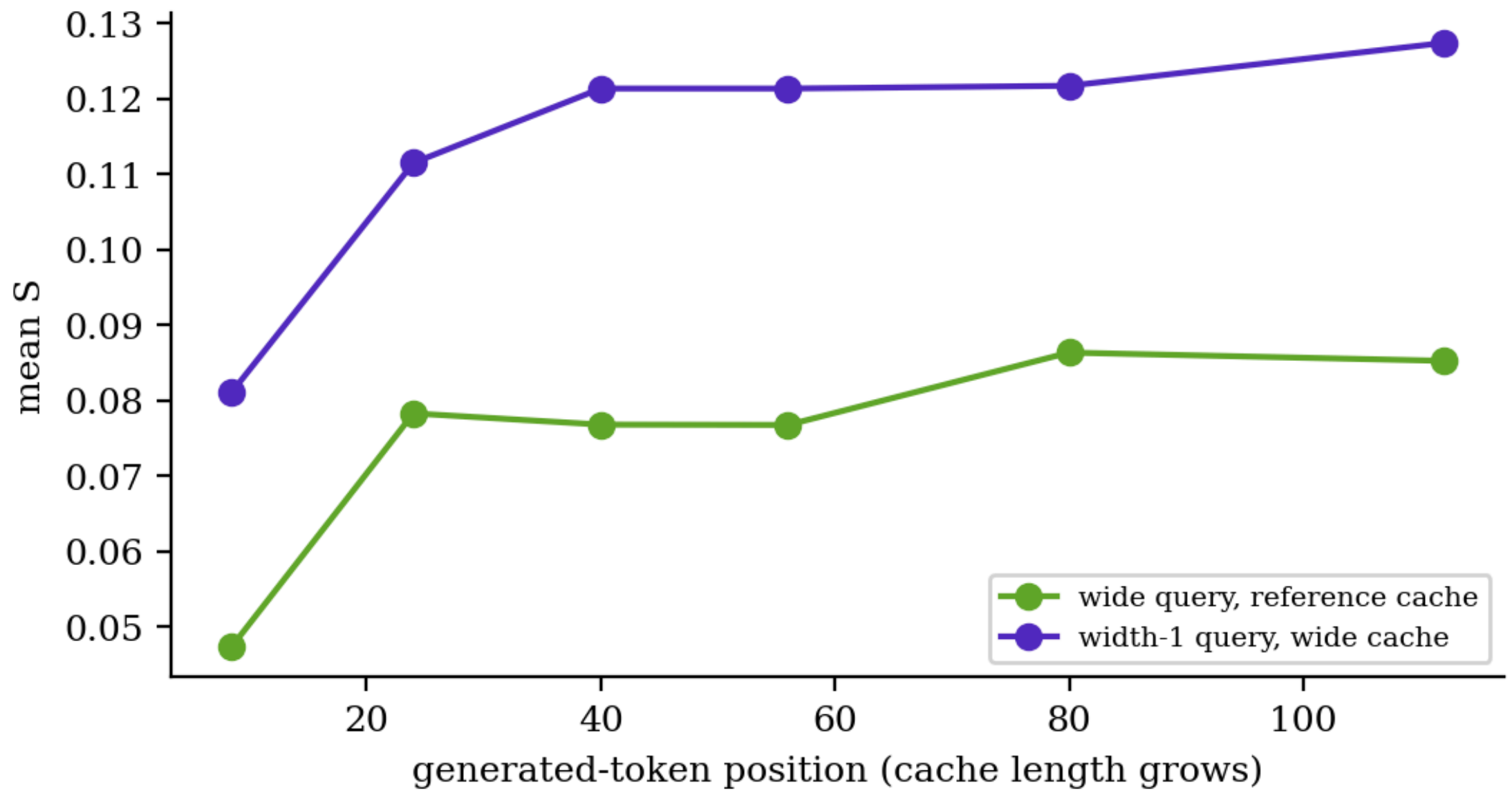

**Figure 13.** Mean noise statistic S by generated position for the query-shape source and the cache source.

## 7 Margin gates and the cache they read

A gate needs a bound and a fallback. The bound comes from the calibration set: eps = q99 = 0.375 is the 99th percentile of S for 4-token verification. When the margin of a verified token is at most eps, the token is not committed from the wide pass, the cache is cut back and a width-1 pass resolves it. The question is which cache that width-1 pass reads. Figure 1 lists the options, and Table 4 and Figure 14 give the result for each.

**Table 4.** Gate variants against the width-1 reference in bfloat16 (eps = q99 unless stated). Variants ran on different numbers of prompts; the comparison with plain verification uses the same prompts. McNemar p-values are exact and not adjusted for the number of variants. Target passes include fallback and rebuild passes.

| variant | prompts | differ from the reference [95% Wilson] (count) | plain verification, same prompts | only variant / only plain | target passes per token | fallbacks per prompt | width-1 fallback passes per prompt |
|---|---|---|---|---|---|---|---|
| plain verification | 150 | 60.0% [52.0, 67.5] (90) | 60.0% (90) | - | 0.75 | 0.00 | 0.0 |

| variant | prompts | differ from the reference [95% Wilson] (count) | plain verification, same prompts | only variant / only plain | target passes per token | fallbacks per prompt | width-1 fallback passes per prompt |
|---|---|---|---|---|---|---|---|
| gate stale q99 | 100 | 58.0% [48.2, 67.2] (58) | 61.0% (61) | 9 / 12, p = 0.664 | 0.86 | 13.55 | 13.6 |
| gate chunk q99 | 60 | 53.3% [40.9, 65.4] (32) | 58.3% (35) | 5 / 8, p = 0.581 | 0.88 | 12.82 | 16.8 |
| gate rebuild-16 q99 | 60 | 31.7% [21.3, 44.2] (19) | 58.3% (35) | 6 / 22, p = 0.004 | 2.37 | 12.90 | 201.9 |
| gate rebuild q99 | 100 | 0.0% [0.0, 3.7] (0) | 61.0% (61) | 0 / 61, p < 0.001 | 7.24 | 13.32 | 808.9 |
| gate rebuild q999 | 45 | 0.0% [0.0, 7.9] (0) | 60.0% (27) | 0 / 27, p < 0.001 | 9.29 | 18.33 | 1065.5 |
| gate rebuild eps 0.25 | 45 | 0.0% [0.0, 7.9] (0) | 60.0% (27) | 0 / 27, p < 0.001 | 4.90 | 9.02 | 517.2 |
| gate rebuild eps 0.125 | 100 | 0.0% [0.0, 3.7] (0) | 61.0% (61) | 0 / 61, p < 0.001 | 3.55 | 5.75 | 350.1 |
| gate rebuild eps 0 | 45 | 33.3% [21.4, 47.9] (15) | 60.0% (27) | 1 / 13, p = 0.002 | 1.59 | 1.78 | 104.1 |
| oracle stale | 100 | 49.0% [39.4, 58.7] (49) | 61.0% (61) | 0 / 12, p < 0.001 | 1.02 | 26.72 | 26.7 |
| oracle rebuild | 30 | 0.0% [0.0, 11.4] (0) | 60.0% (18) | 0 / 18, p < 0.001 | 1.24 | 0.90 | 63.5 |

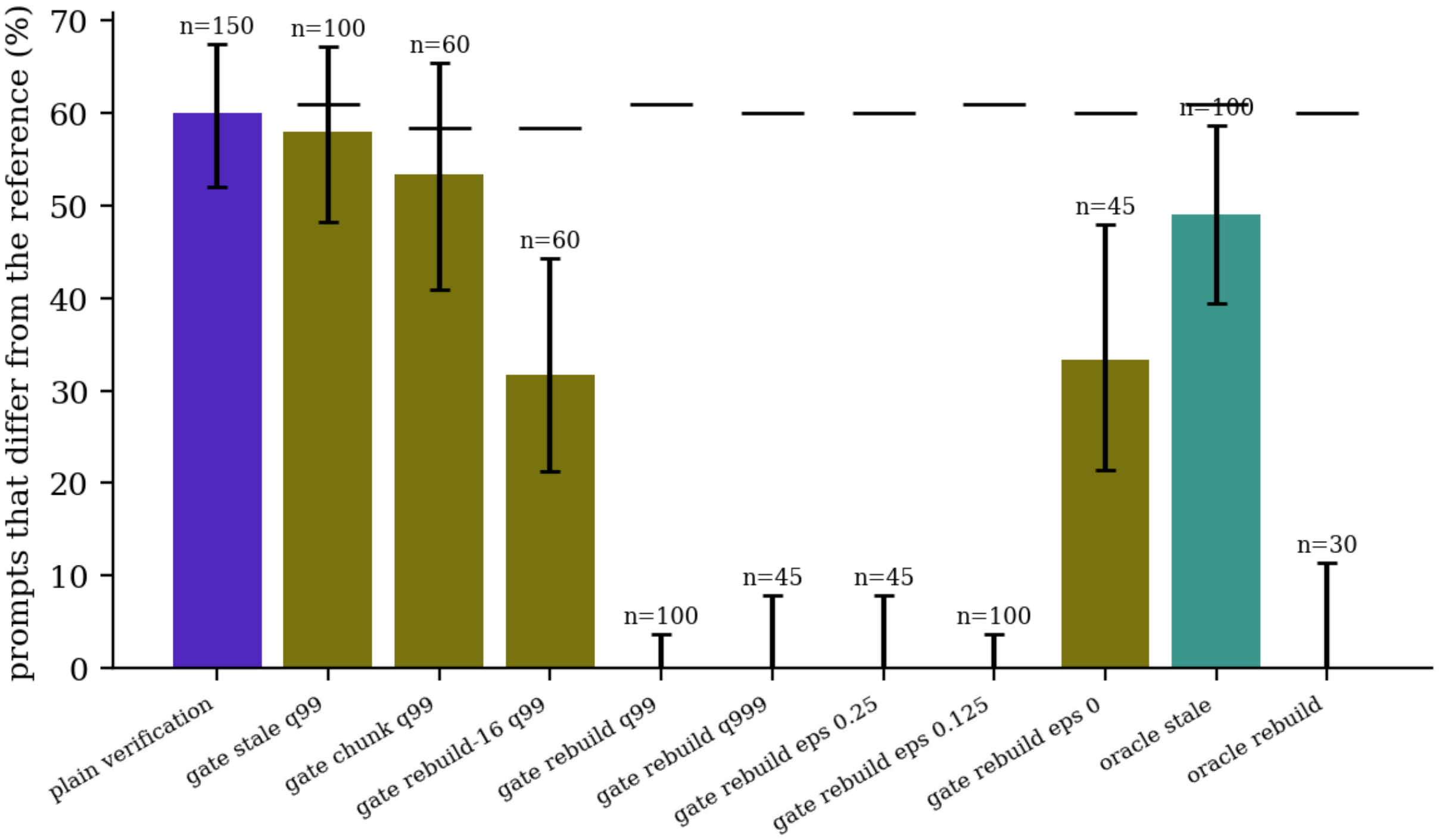

**Figure 14.** Share of prompts that differ from the reference for plain verification and for each fallback variant (bars, 95% Wilson intervals). The black tick marks plain 4-token verification on the same prompts. Oracle variants fall back exactly where the verify argmax differs from the reference.

With a stale cache the gate does not help: 58.0% [48.2, 67.2] (58/100) of prompts still differ, against 61/100 for plain verification on the same prompts (McNemar $p = 0.664$). The fallback pass reads entries that wide passes wrote, and Section 6 showed that such a pass flips the argmax about as often as the wide pass does.

Recomputing only the current chunk before the fallback is not enough either: 53.3% [40.9, 65.4] (32/60) of prompts differ, against 35/60 for plain verification on the same prompts. Entries from earlier chunks remain wide-written.

Rebuilding the whole generated part of the cache with width-1 passes before the fallback removes the divergence: 0.0% [0.0, 3.7] (0/100) of prompts differ, against 61/100 for plain verification on the same prompts (McNemar p = 8.7e-19). The price is high. The rebuild variant spends 7.24 target forward passes per generated token, against 1.00 for the reference itself and 0.75 for plain verification, because it makes 13.3 fallbacks per prompt and 809 width-1 passes in them. The bound flags a large share of positions: 10.7% of generated tokens trigger a fallback. That the rebuild gate reaches zero is expected from the exact replay check in Section 3, because the fallback then reads the reference cache. What the experiment adds is the comparison with the other fallbacks and the cost.

The oracle shows that the trigger is not the limit. A stale-cache gate that falls back exactly where the verify argmax differs from the reference still leaves 49.0% [39.4, 58.7] (49/100) of prompts different, against 61/100 for plain verification on the same prompts. Even a perfect trigger cannot repair a width-1 pass that reads wide-written cache entries, and this is the cleanest evidence we have for cache noise as the cause.

The oracle with a rebuilt cache differs from the reference on 0.0% [0.0, 11.4] (0/30) of prompts, as it must if the replay is exact, and it costs 1.24 target passes per token with 0.9 fallbacks per prompt.

Rebuilding only the last 16 cache entries costs 2.37 target passes per token (202 fallback passes per prompt) and leaves 31.7% [21.3, 44.2] (19/60) of prompts different, against 35/60 for plain verification on the same prompts.

A looser bound (eps = q999 = 0.625) with the full rebuild leaves 0.0% [0.0, 7.9] (0/45) of prompts different at 9.29 target passes per token and 18.3 fallbacks per prompt.

The bound sets the cost. A larger eps sends more positions to the fallback, so shrinking it saves rebuilds (Figure 15). The smallest bound that left no divergence is eps = 0.125, one step of the bfloat16 grid at these logit sizes. It sends 5.8 positions per prompt to the fallback, spends 3.55 target passes per token and leaves 0.0% [0.0, 3.7] (0/100) of prompts different, against 61.0% for plain verification on the same prompts. The calibrated q99 bound (0.375) is exact as well but sends 13.3 positions per prompt to the fallback and spends 7.24 passes per token, so it is conservative by a factor of about 2.0 in cost. At eps = 0.000 the gate leaves 33.3% [21.4, 47.9] (15/45) of prompts different, at 1.59 passes per token. Even the cheapest exact bound costs more passes than the reference.

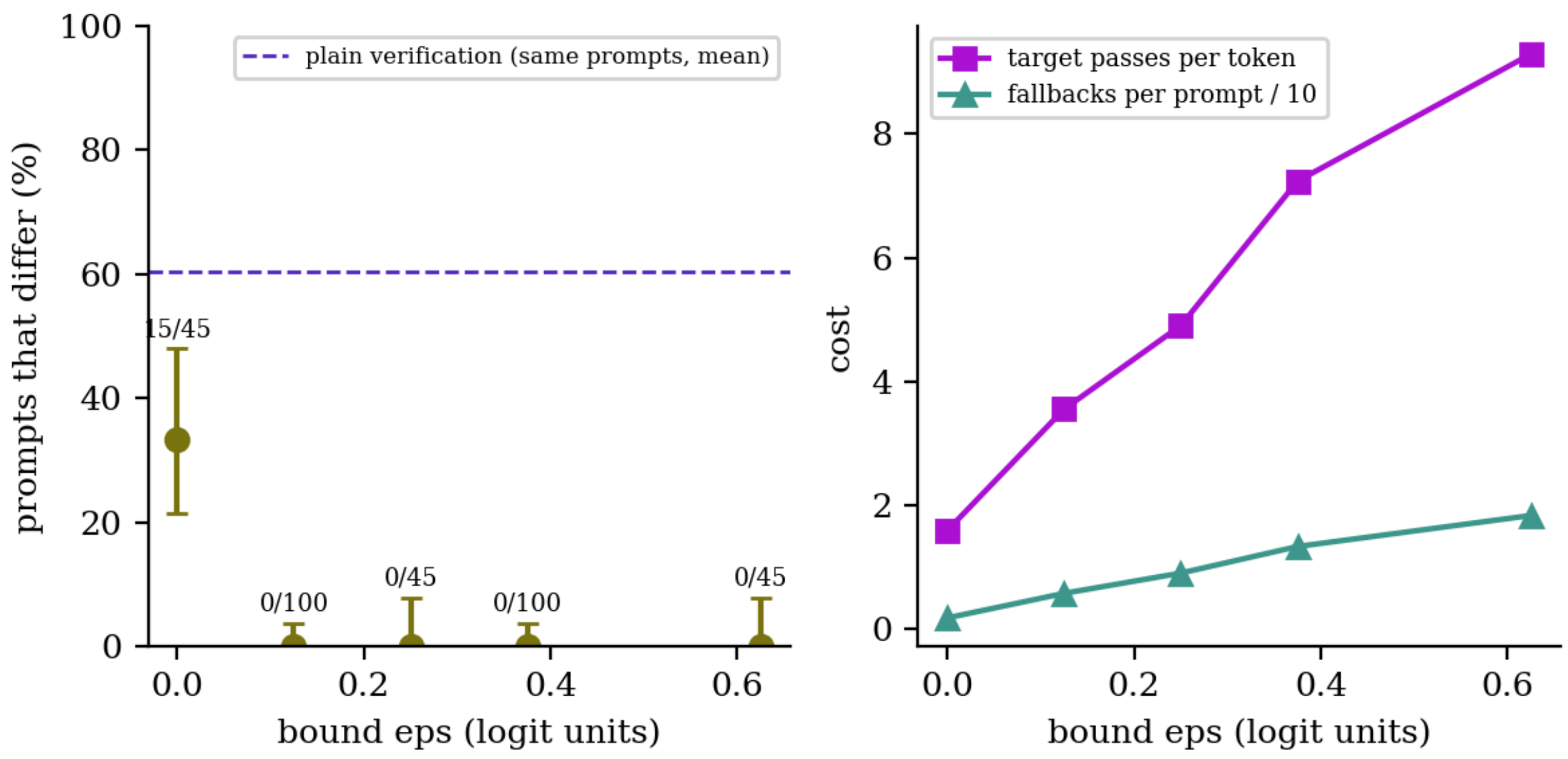

**Figure 15.** Rebuild gate as the bound eps shrinks. Left: share of prompts that differ from the reference (95% Wilson intervals; counts above the points). Right: target passes per token and fallbacks per prompt (divided by 10).

The variants form a dose-response (Figure 16): the more of the cache the fallback recomputes with width-1 passes, the fewer prompts differ. On the 60 prompts that all four variants finished, the divergence share falls from the stale cache through the current chunk and the last 16 entries to all generated entries, and the last variant reaches zero. A cache-noise explanation predicts this pattern. A fallback-trigger explanation does not.

Figure 17 puts exactness against cost. Every variant that reaches zero divergences spends more target forward passes per token than the reference does, and the fallbacks are frequent: Figure 18 shows the counts. The cost of the rebuild gate is dominated by the width-1 passes in the rebuild. With an oracle trigger the rebuild gate also reaches zero differences, at 1.24 target passes per token, so a better trigger would cut the cost, but it stays above the reference. The dose-response suggests that entries older than 16 positions matter. If every committed cache entry must be rewritten by a width-1 pass for the cache to match, the cost cannot fall below that of the reference, which spends one such pass per token; we did not prove that necessity.

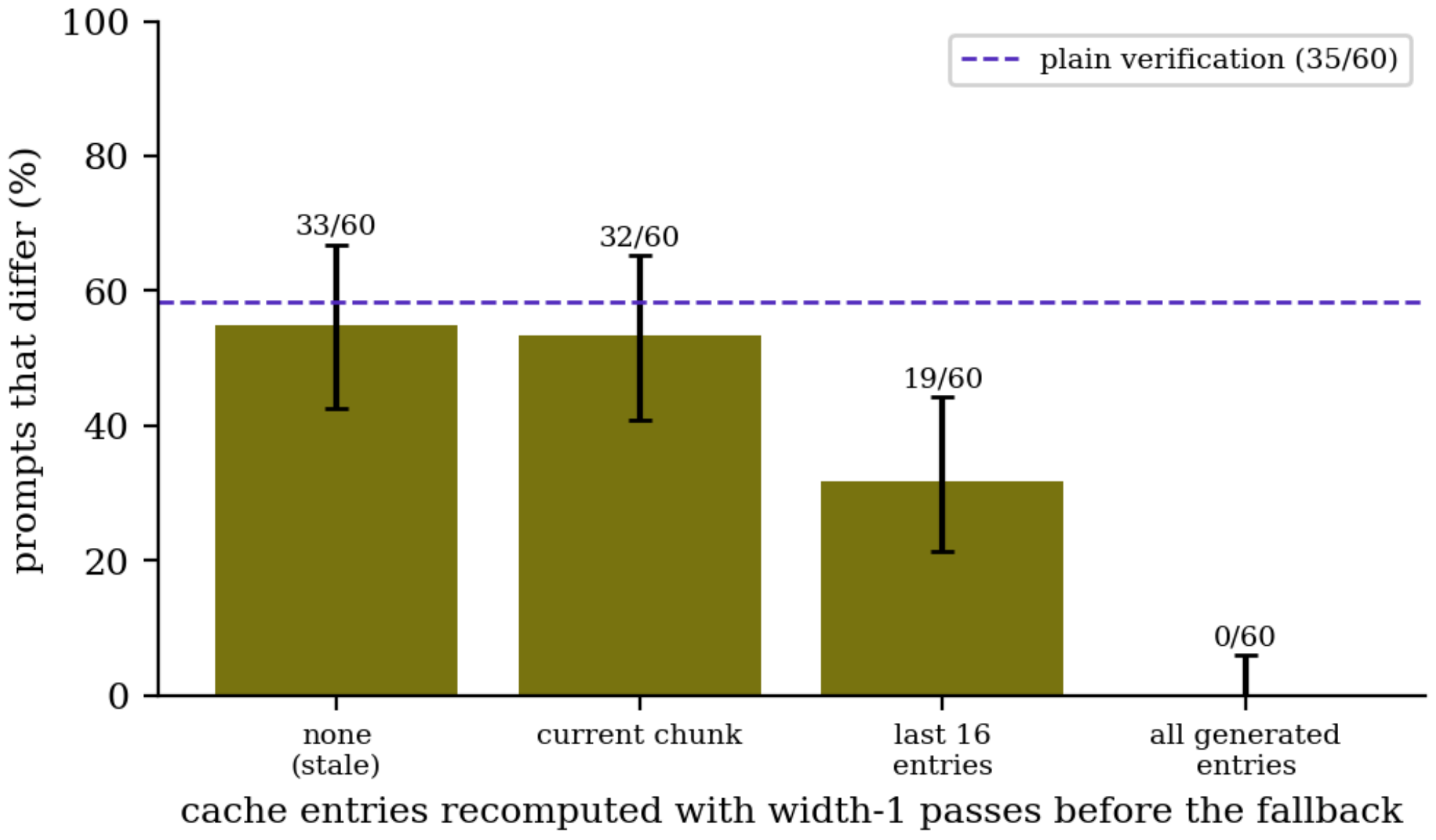

**Figure 16.** Dose-response of the fallback: divergence share on the 60 prompts that every variant finished, by how much of the cache is recomputed with width-1 passes before the width-1 fallback pass (eps = q99 for all). 95% Wilson intervals.

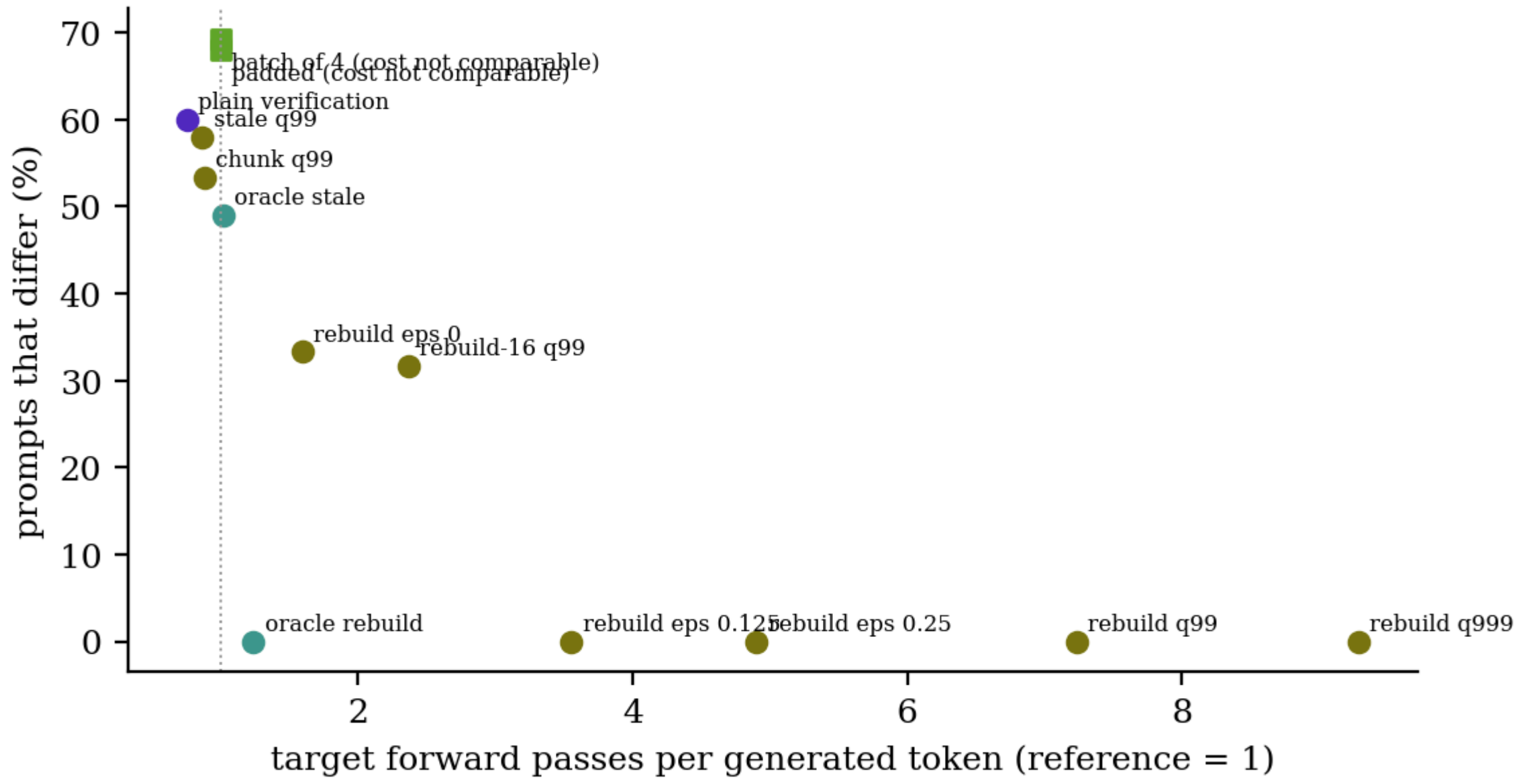

**Figure 17.** Exactness against cost. Forward passes include the width-1 fallback and rebuild passes. Fewer passes per token is cheaper in the memory-bound regime; the reference costs 1.0. Square markers are paths that do not use drafts.

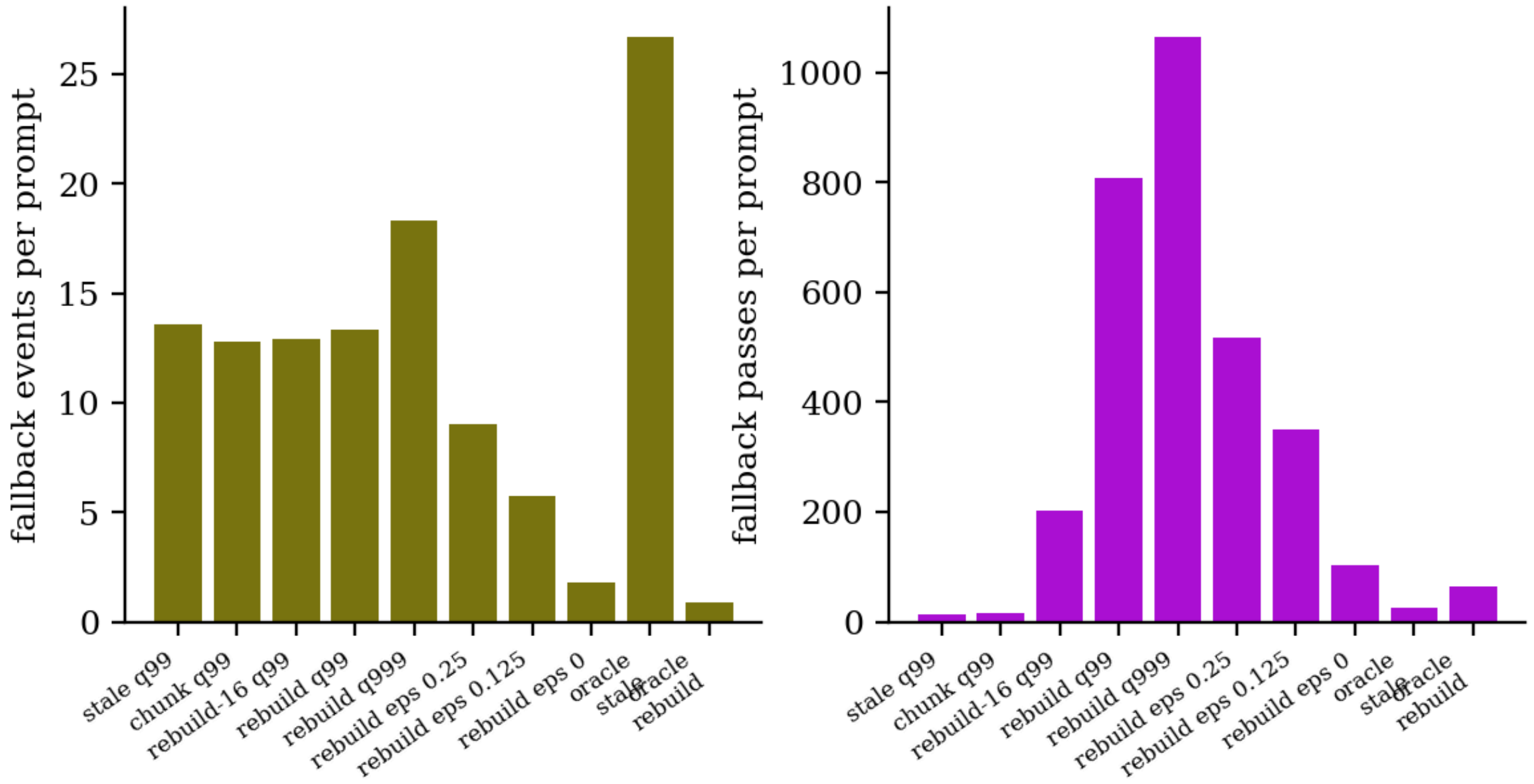

**Figure 18.** Fallback events and width-1 passes per prompt for each variant. Rebuilding the cache costs more width-1 passes per event because it recomputes the generated part of the cache.

Wall-clock time is context only, because the GPU was shared and each prompt was timed once in one process on the first 12 evaluation prompts. Table 5 gives milliseconds per generated token and the ratio to the reference. Fixed-width decoding with drafts runs at 0.68 of the reference time per token [0.66, 0.71], and the rebuild gate at 6.13 [4.45, 7.88]. A ratio below 1 means faster than the reference. These ratios depend on this Python decode loop, on the small model size and on the neighbours of a shared GPU, so we draw no speed claim from them beyond the order of magnitude.

**Table 5.** Wall-clock context on a shared GPU: median milliseconds per generated token over prompts, and the mean ratio to the reference with a bootstrap interval over prompts. Not a speed benchmark.

| path | prompts | median ms per token | mean ratio to the reference [95% bootstrap] |
|---|---:|---:|---:|
| reference | 12 | 26.9 | 1.00 [1.00, 1.00] |
| 4-token verification | 12 | 18.7 | 0.71 [0.67, 0.75] |
| fixed width 5, plain | 12 | 25.7 | 0.95 [0.93, 0.96] |
| fixed width 5, with drafts | 12 | 18.0 | 0.68 [0.66, 0.71] |
| gate stale q99 | 12 | 21.5 | 0.81 [0.75, 0.87] |
| gate rebuild q99 | 12 | 127.5 | 6.13 [4.45, 7.88] |
| gate rebuild eps 0.125 | 12 | 85.6 | 3.44 [2.49, 4.38] |

## 8 Fixed-shape decoding and the width scan

If exactness against a width-1 reference costs more than the reference, the alternative is to change what the reference is. Batch-invariant kernels do this by giving every request the same reduction order (arXiv:2601.17768, arXiv:2605.14220). The analogue here is fixed-shape decoding. Every pass after the prefill has the same width $W = 5$. Proposed tokens fill the rows when there are any, and a filler token fills the rest, so a speculative step and a plain step run the same kernels over the same shapes. If rounding depends only on the pass shape, plain decoding at width 5 and decoding with drafts at width 5 must agree exactly, whatever the drafts. We ran both on 100 prompts (Figure 19).

Fixed-width decoding with drafts and plain fixed-width decoding differ on 0.0% [0.0, 3.7] (0/100) of prompts. Both differ from the width-1 reference (66.0% [56.3, 74.5] (66/100) for drafts, 66.0% [56.3, 74.5] (66/100) for plain), as variable-width verification does (61.0% [51.2, 70.0] (61/100) on the same prompts).

Fixed-width decoding with drafts costs 0.75 target passes per token, the same as variable-width verification (0.75), and plain fixed-width decoding costs 1.00. So fixed-shape decoding makes speculation consistent with itself and not with the width-1 path: it buys reproducibility across draft choices at no extra passes, and it does not buy agreement with a path that was never fixed. A pass of width 5 does more arithmetic than a pass of width 1, which the pass counts do not show.

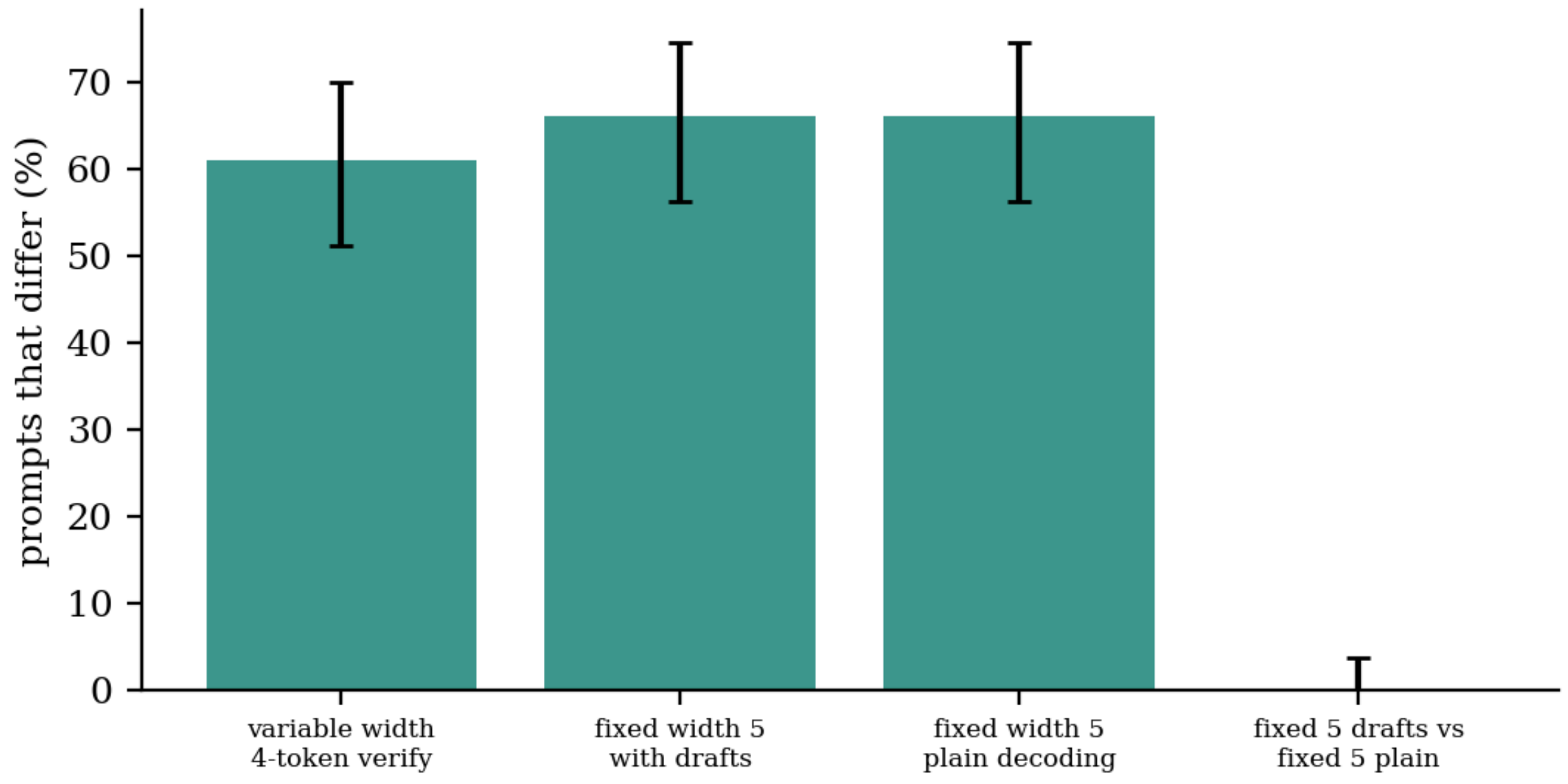

**Figure 19.** Fixed-shape decoding (100 prompts). The first three bars compare with the width-1 reference; the last compares fixed-width decoding with drafts against fixed-width decoding without drafts.

The width scan (Figure 20) shows why. Teacher-forced along the reference tokens of 20 prompts (2459 positions), we compare the whole logit row (every vocabulary entry) of a pass of width W with the row of another width. A width-1 pass differs from every wider pass: at most 0.0% of positions have a wider row that is bit-identical to the width-1 row, for any width from 2 to 32. Widths 2 and 3 produce identical rows (100.0% of positions), and widths 4, 5 and 6 produce identical rows (100.0% of positions for width 4 and 100.0% for width 6, each against width 5). The width-2 row equals the width-5 row at 76.5% of positions. Width 8 matches width 5 at 65.4% of positions, and widths 12, 16, 32 match width 5 at 0.0% of positions or fewer. Widths therefore fall into groups that share a kernel, and a width-1 pass shares none with a wider pass. That is the numerical reason for everything above: the reference is the only path whose cache and logits come from the single-row kernel, and variable-width verification mixes the group of widths 2 and 3 with the group of widths 4 to 6.

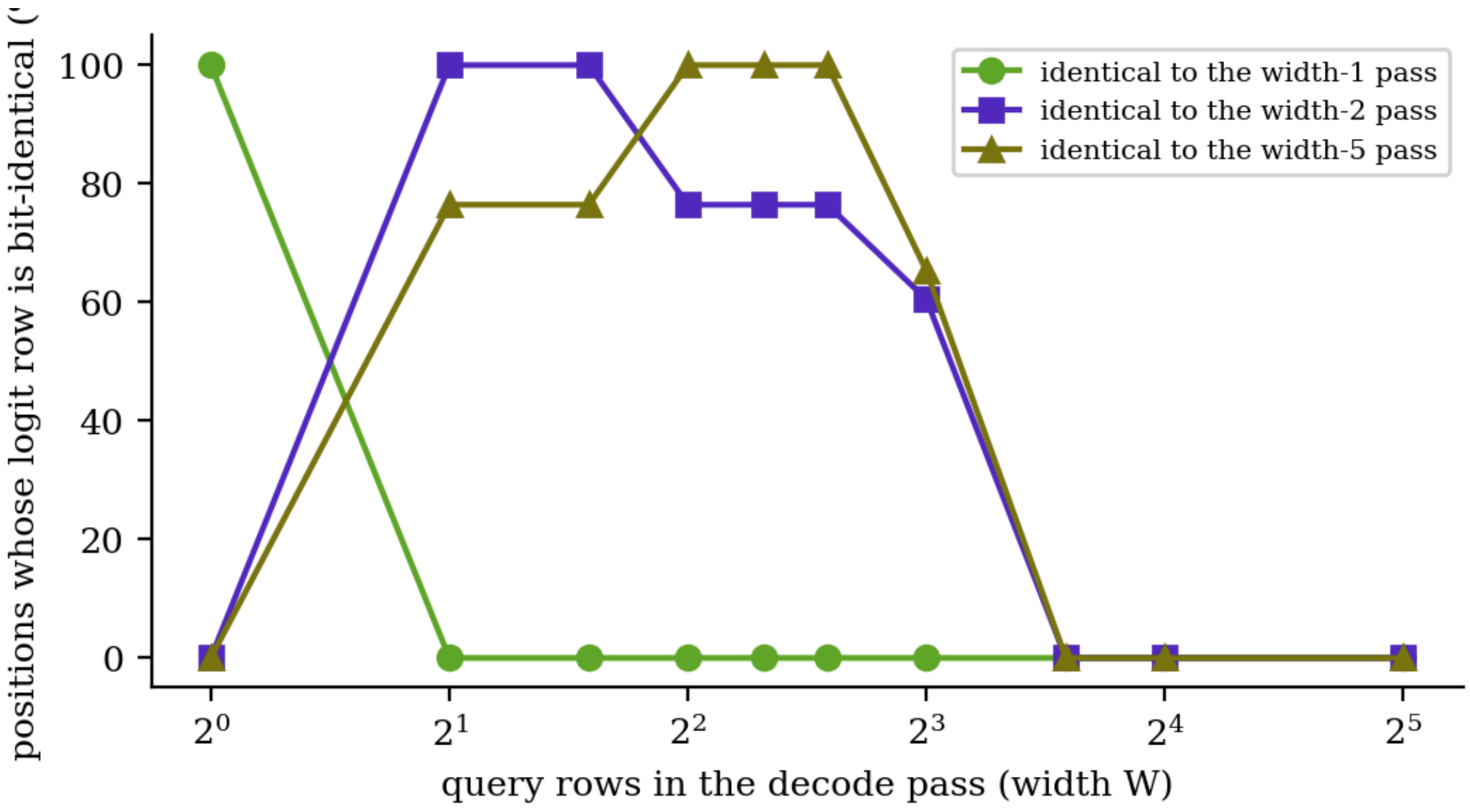

**Figure 20.** Bit-level agreement of the whole logit row between decode passes of width W and the width-1 and width-2 passes, teacher-forced along the reference tokens of 20 prompts (2459 positions).

## 9 Does the bound transfer to prompts from another distribution?

The bound was set on GSM8K prompts only (Section 3), so the ARC evaluation prompts are already a transfer test. Split by family, the rebuild gate leaves 0.0% [0.0, 7.1] (0/50) of GSM8K prompts and 0.0% [0.0, 7.1] (0/50) of ARC prompts different, while the stale-cache gate leaves 42.0% [29.4, 55.8] (21/50) and 74.0% [60.4, 84.1] (37/50) and plain verification 41.1% [30.5, 52.6] (30/73) and 77.9% [67.5, 85.7] (60/77).

The shifted set (MMLU and HellaSwag prompts) changes the topics, the prompt format and the answer style. On it, 4-token verification changes 80.0% [68.2, 88.2] (48/60) of 60 prompts, and the reference margin is at most 0.25 at 10.5% of positions, against 8.1% on the evaluation prompts (Figure 21).

The stale-cache gate with the calibrated bound (eps = q99 = 0.375) leaves 68.3% [55.8, 78.7] (41/60) of shifted prompts different, against 48/60 for plain verification on the same prompts, at 0.90 target passes per token.

The rebuild gate with the calibrated bound (eps = q99 = 0.375) leaves 0.0% [0.0, 6.0] (0/60) of shifted prompts different, against 48/60 for plain verification on the same prompts, at 10.71 target passes per token.

The bound also covers the divergences themselves (Figure 22): the verify-pass margin at the first divergence of plain verification is at most q99 for 100.0% [95.9, 100.0] (90/90) on the evaluation prompts; 100.0% [92.6, 100.0] (48/48) on the shifted prompts.

**Table 6.** Shifted prompts (MMLU and HellaSwag): divergence from the width-1 reference by variant, with the bound calibrated on the calibration prompts.

| variant | prompts | differ from the reference [95% Wilson] (count) | plain verification, same prompts | only variant / only plain | target passes per token | fallbacks per prompt | width-1 fallback passes per prompt |
|---|---|---|---|---|---|---|---|
| plain verification | 60 | 80.0% [68.2, 88.2] (48) | 80.0% (48) | - | 0.76 | 0.00 | 0.0 |
| gate stale q99 | 60 | 68.3% [55.8, 78.7] (41) | 80.0% (48) | 3 / 10, p = 0.092 | 0.90 | 17.32 | 17.3 |
| gate rebuild q99 | 60 | 0.0% [0.0, 6.0] (0) | 80.0% (48) | 0 / 48, p < 0.001 | 10.71 | 18.03 | 1264.3 |

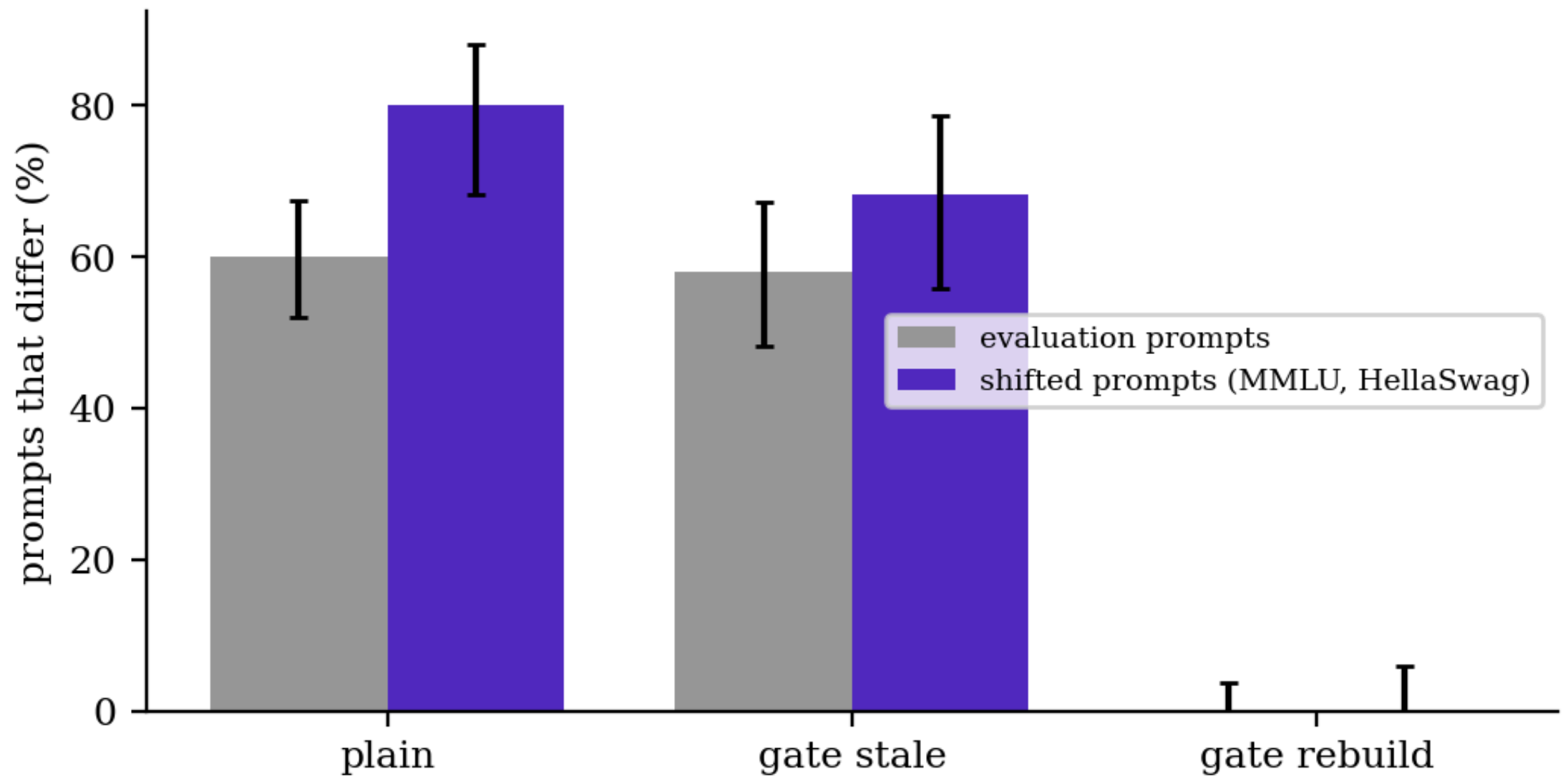

**Figure 21.** Held-out shift: the noise bound is calibrated on the GSM8K and ARC calibration prompts; the shifted prompts come from MMLU and HellaSwag. Bars show 95% Wilson intervals.

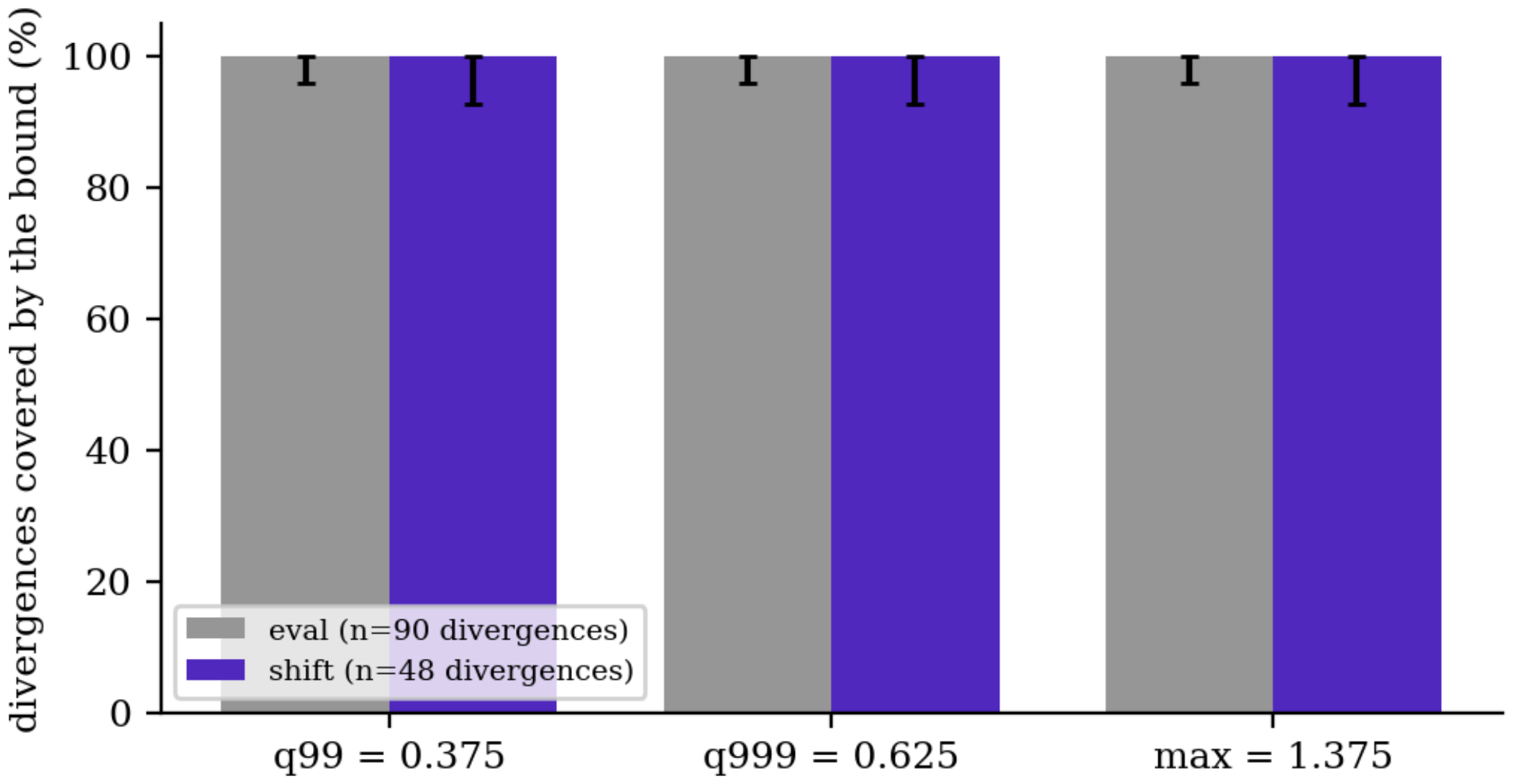

**Figure 22.** Coverage of first divergences of plain verification by the calibrated margin bound: the share whose verify-pass margin is at most the bound, on the evaluation and the shifted prompts.

The bound can also be tested without generating text, by teacher forcing. On prompts that did not set it, the share of positions with S above q99 = 0.375 should be about 1% if the calibration prompts are exchangeable with them (Figure 23, Table 7). It is 37 of 5253 positions (0.70%) on the calibration prompts (bound set here) (42 prompts), and 53 of 9626 positions (0.55%) on the further calibration-set prompts (78 prompts), and 46 of 7552 positions (0.61%) on the shifted prompts (MMLU, HellaSwag) (60 prompts). A gate at q99 misses a flip when the verify margin lies above the bound, and the numbers of flips missed are 0 of 37 on the calibration prompts (bound set here); 0 of 162 on the further calibration-set prompts; 0 of 147 on the shifted prompts (MMLU, HellaSwag).

**Table 7.** Coverage of the noise bound on prompts that did not set it. S is for 4-token verification (wide query, wide cache). Columns: prompts, positions, positions with S above q99 (share, 95% Wilson interval), above q999, above the calibration maximum, argmax flips, flips whose verify margin lies above the q99 bound (missed by a gate), and the share of positions a gate at q99 flags.

| set | prompts | positions | S above q99 | S above q999 | S above max | flips | flips missed | positions flagged |
|---|---|---|---|---|---|---|---|---|

| set | prompts | positions | S above q99 | S above q999 | S above max | flips | flips missed | positions flagged |
|---|---|---|---|---|---|---|---|---|
| calibration prompts (bound set here) | 42 | 5253 | 37 (0.70%) [0.51, 0.97] | 4 (0.08%) | 0 | 37 | 0 | 6.3% |
| further calibration-set prompts | 78 | 9626 | 53 (0.55%) [0.42, 0.72] | 2 (0.02%) | 0 | 162 | 0 | 11.9% |
| shifted prompts (MMLU, HellaSwag) | 60 | 7552 | 46 (0.61%) [0.46, 0.81] | 4 (0.05%) | 0 | 147 | 0 | 14.2% |

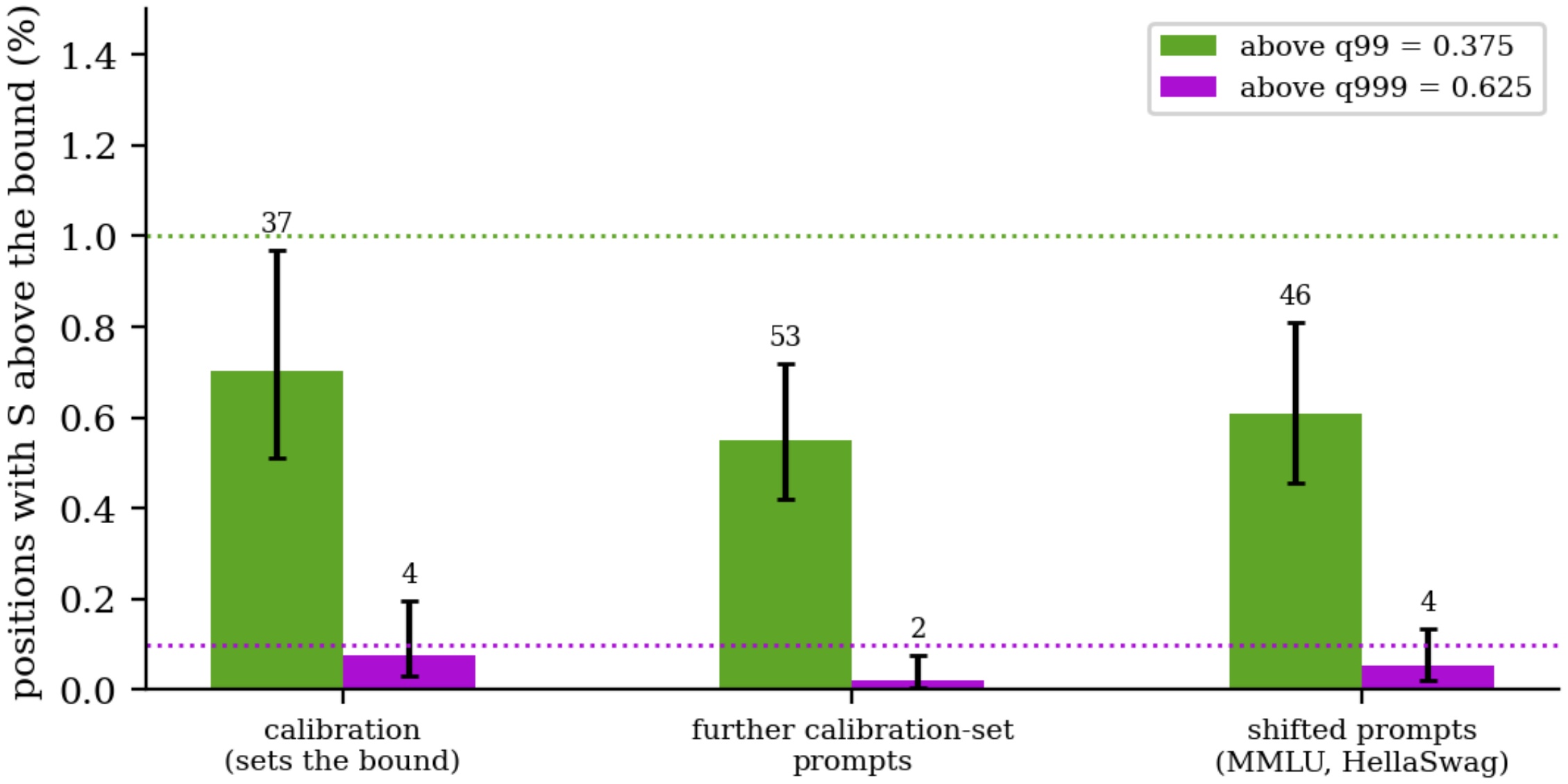

**Figure 23.** Share of positions whose noise statistic S for 4-token verification exceeds the calibrated bound, by prompt set (counts above bars, 95% Wilson intervals). Dotted lines mark the share expected by construction on exchangeable prompts: 1% for q99 and 0.1% for q999.

## 10 How close is the reference to float32?

Exactness against the reference is a statement about reproducibility. Is the reference also more accurate? We ran a float32 copy of the model over the same teacher-forced sequences (43 calibration prompts, 5380 positions) and counted the positions where each bfloat16 path picks a different argmax from float32 (Figure 24). The bfloat16 reference disagrees with float32 at 34 positions (6.3 per 1000), the verify pass at 24 (4.5), the cache-noise pass at 25 and the padded whole-sequence pass at 29. The reference disagrees with float32 at no fewer positions than the verify pass, the cache-noise pass or the padded pass, so agreement with the width-1 path is not agreement with a more exact computation.

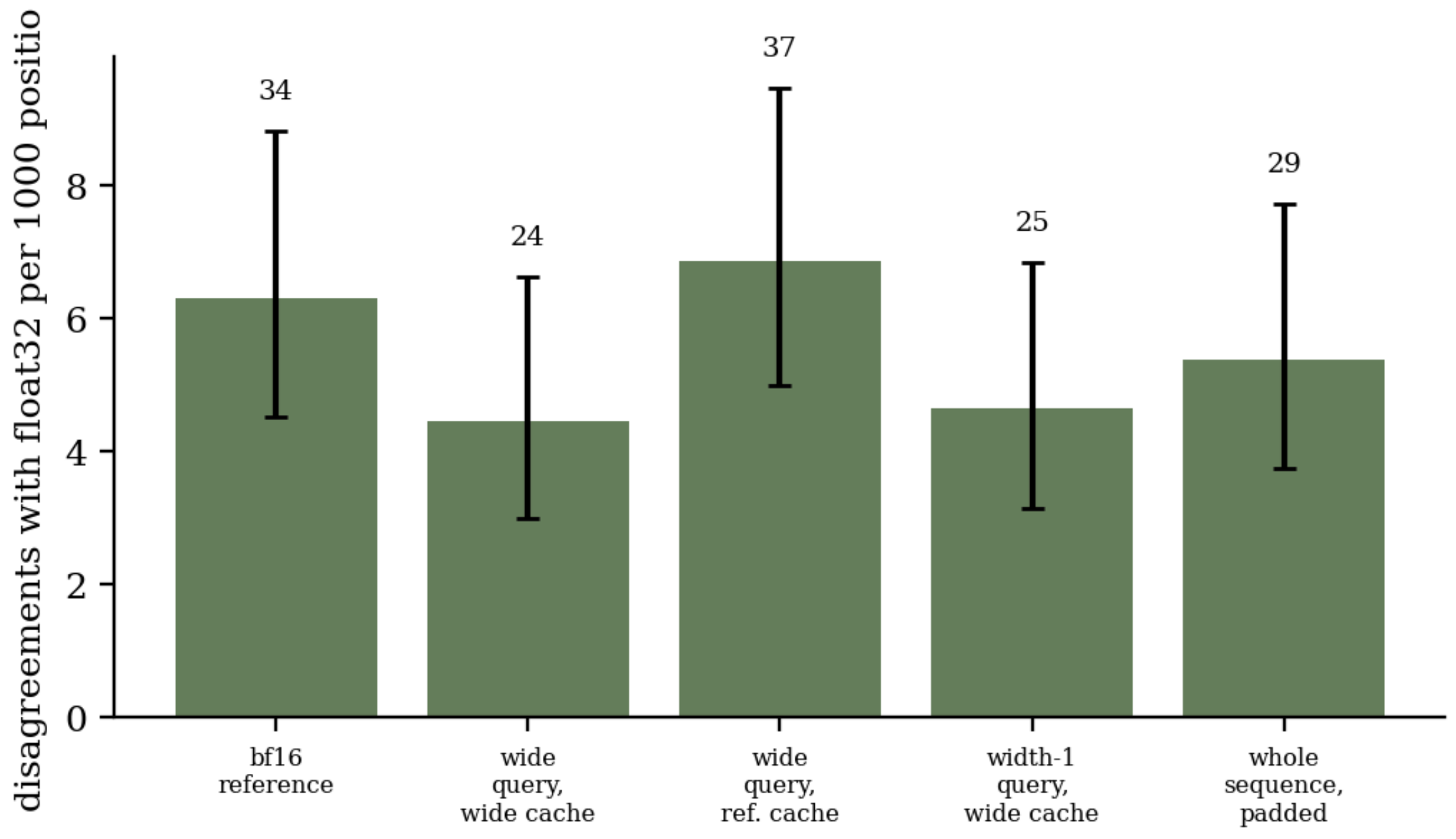

**Figure 24.** Disagreement of each bfloat16 path with the float32 argmax along the reference tokens (43 calibration prompts, 5380 positions). Counts above the bars.

## 11 Cheap paths and a second model size

The same margin signal can gate approximations that are deliberately inexact. Table 8 reports, along the reference tokens of 100 evaluation prompts (12407 positions), how often a cheaper path picks the reference argmax. The first kind is a logit-lens read-out at an earlier layer, which is the exit used by early-exit decoding (arXiv:2207.07061) applied here without a trained exit head. The second is simulated round-to-nearest weights in groups of 128, with per-group symmetric scales dequantised back to bfloat16, so the arithmetic is simulated. The last two columns restrict to positions where the cheap path's own margin is at least 2 logit units. They give the share of positions kept and the agreement among them (Figure 25).

**Table 8.** Cheap paths against the bfloat16 full model along the reference tokens: agreement of the argmax overall, and among positions where the cheap path's own margin is at least 2.

| path | agree with the full model | positions with margin at least 2 | agree among those |
|---|---:|---:|---:|
| exit at layer 7 of 28 | 1.6% | 14.7% | 2.6% |
| exit at layer 9 of 28 | 3.3% | 15.0% | 4.3% |
| exit at layer 14 of 28 | 4.8% | 16.5% | 8.5% |
| exit at layer 17 of 28 | 8.9% | 22.8% | 15.1% |
| exit at layer 21 of 28 | 42.6% | 48.6% | 60.6% |
| exit at layer 24 of 28 | 67.6% | 66.6% | 82.1% |
| exit at layer 26 of 28 | 82.3% | 71.7% | 93.5% |
| exit at layer 27 of 28 | 85.9% | 67.1% | 97.0% |
| full depth, whole-sequence pass | 99.1% | 63.8% | 100.0% |
| 8-bit round-to-nearest weights | 98.4% | 63.6% | 100.0% |
| 6-bit round-to-nearest weights | 95.0% | 64.1% | 99.9% |
| 4-bit round-to-nearest weights | 80.9% | 60.2% | 95.7% |
| 3-bit round-to-nearest weights | 21.6% | 15.8% | 54.6% |

Gating by margin raises agreement for every cheap path, as it does for the rounding noise above. Exiting at layer 26 agrees at 82.3% to 93.5% while keeping 71.7% of positions, and exiting at layer 27 agrees at 85.9% to 97.0% while keeping 67.1% of positions. For the weights, 8 bits go from 98.4% to 100.0% while keeping 63.6% of positions, 4 bits from 80.9% to 95.7% while keeping 60.2% of positions and 3 bits from 21.6% to 54.6% while keeping 15.8% of positions. Early exits well before the last layers agree too rarely for a margin gate to rescue them without a trained read-out. The full-depth row is the whole-sequence pass of the same model and shows the rounding floor on these positions, at 99.1% agreement.

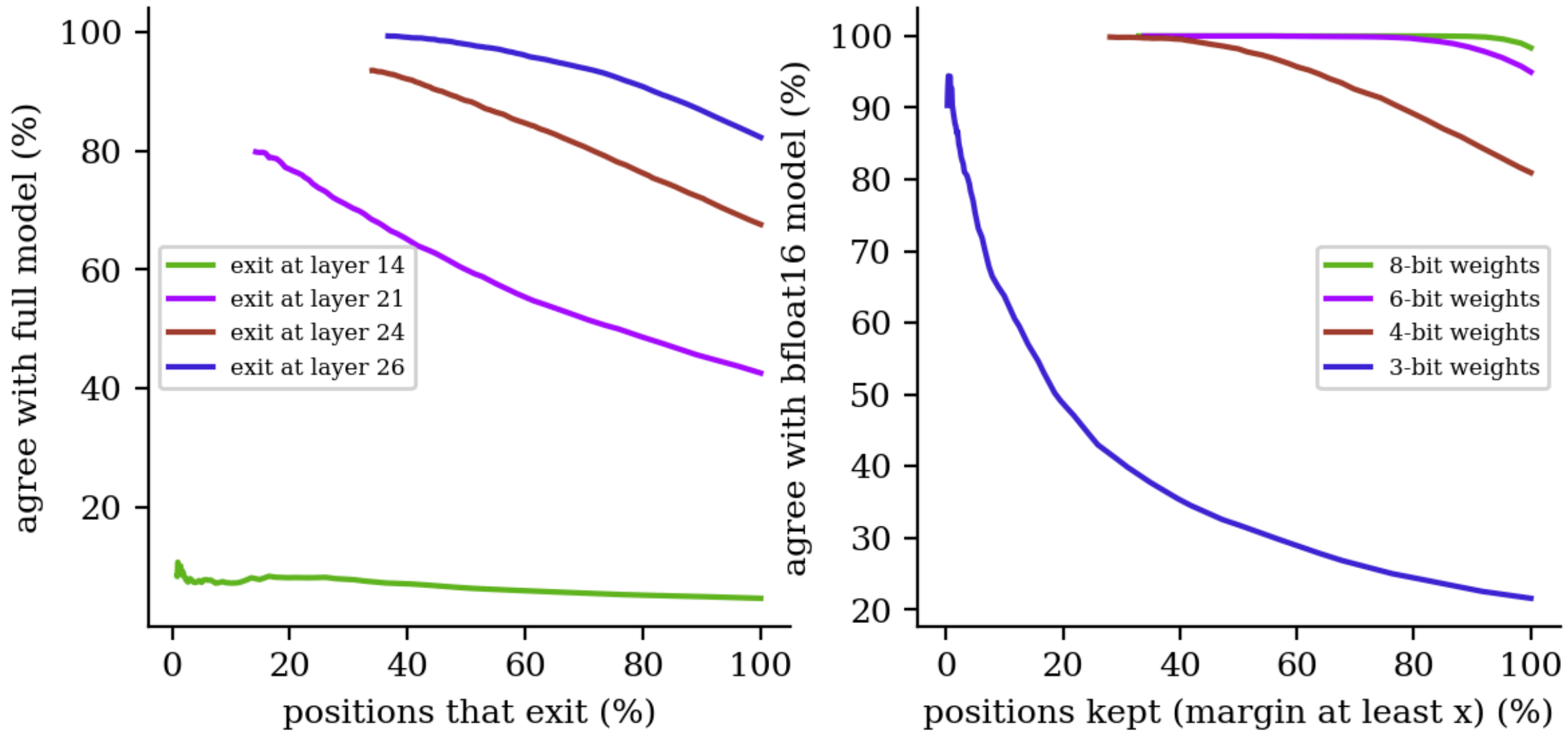

**Figure 25.** Cheap paths gated by margin: logit-lens early exit at several layers (left) and round-to-nearest weights (right), teacher-forced along the reference tokens. Sweeping the margin threshold trades the share of positions that take the cheap path against its agreement with the full model.

A second model size checks whether the rates are specific to 0.6B parameters. On the 45 evaluation prompts that both models finished, 4-token verification changes 60.0% [45.5, 73.0] (27/45) for Qwen3-0.6B and 68.9% [54.3, 80.5] (31/45) for Qwen3-1.7B, and padding changes 71.1% [56.6, 82.3] (32/45) and 73.3% [59.0, 84.0] (33/45) (Figure 26). The 95% intervals of the two sizes overlap for both paths, so we do not rank the sizes.

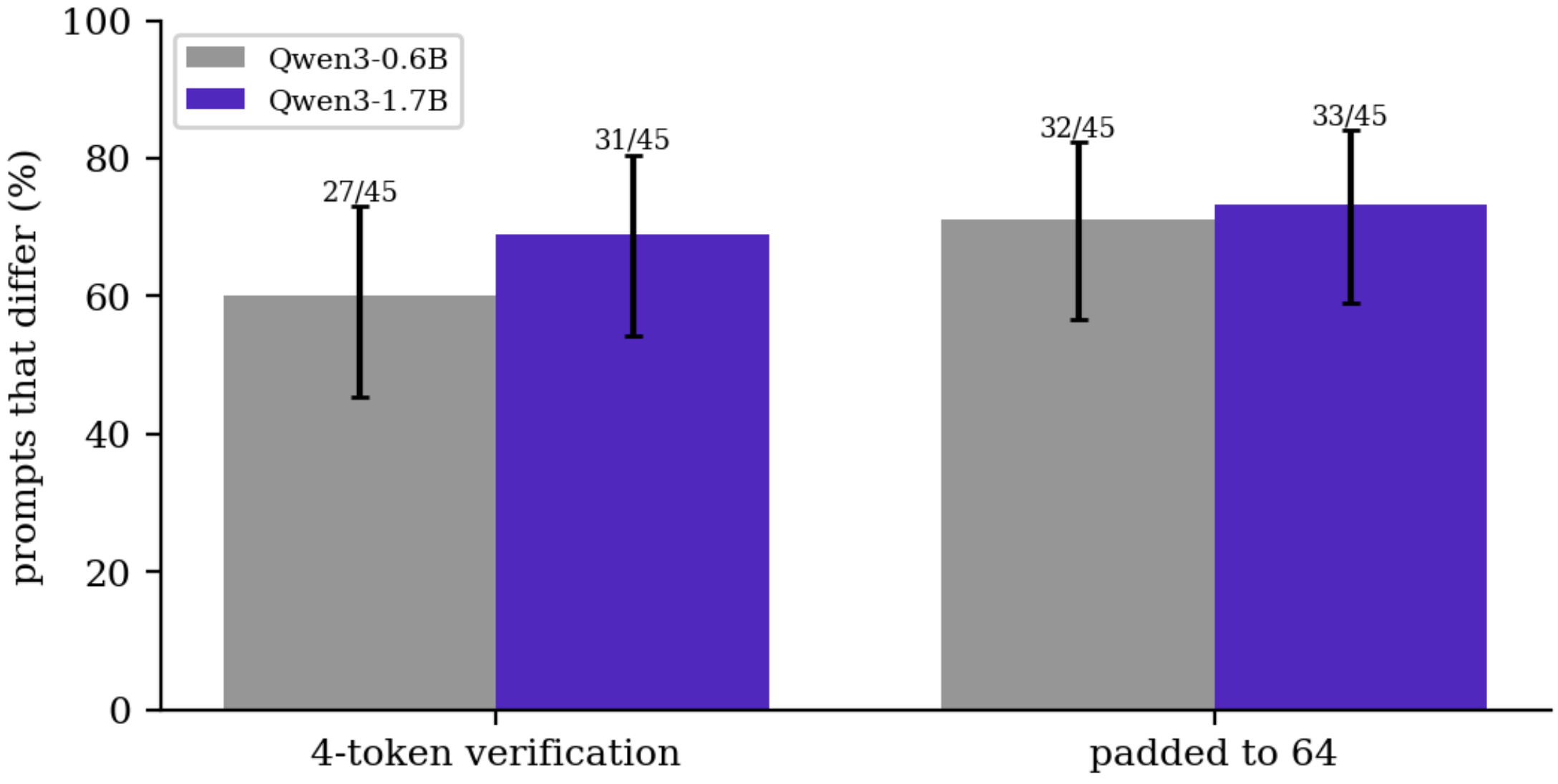

**Figure 26.** Divergence rate by model size on the evaluation prompts that both models finished for each path, 95% Wilson intervals (counts above bars).

The noise split repeats on Qwen3-1.7B. On 40 calibration prompts (5080 positions) the mean of S is 0.157 for the cache source, 0.102 for the query-shape source and 0.158 for both together, and the argmax flips are 36 for the cache source, 34 for the query-shape source and 33 for both together. For the 0.6B model the corresponding means are 0.117, 0.078 and 0.120. The 99th percentile of S for the larger model is 0.500, against 0.375 for the smaller one.

The gates also carry over to the larger model with the bound taken unchanged from the 0.6B calibration (eps = 0.375), which tests the bound across model sizes. The stale-cache gate leaves 66.7% [52.1, 78.6] (30/45) of prompts different, against 31/45 for plain verification on the same prompts, at 0.80 target passes per token. The rebuild gate leaves 0.0% [0.0, 7.9] (0/45) of prompts different, against 31/45 for plain verification on the same prompts, at 3.28 target passes per token.

## 12 Does a divergence change the answer?

A divergence is a different token sequence, which need not be a different answer. For the ARC prompts we extract the first stated option letter from each output with a regular expression ("correct answer" or "answer" followed by a letter A to D). This works for most ARC outputs and fails on others, so the comparison uses the prompts where both outputs state a letter. GSM8K outputs are cut at 128 tokens before the final number in most cases, so we do not compare them. Table 9 lists, per path, the ARC prompts whose output differs from the reference, the differing outputs where both state a letter, how many of those letters changed, and the accuracy of the reference and of the path against the gold letter on the prompts where both state one.

Of the 60 ARC prompts where 4-token verification changed the text, 52 state a letter in both outputs, and the letter changed in 0 of them. For padding the figures are 64, 53 and 2. Pooled over the paths that diverge, the letter changed in 2 of 192 differing prompt-path pairs with both letters (1.0% [0.3, 3.7]). The sample is a few dozen prompts per path, so the rate of letter changes is poorly bounded. Most divergences in this sample therefore change the wording of the explanation and not the option that the model names. The accuracy columns are descriptive (few prompts, crude extraction) and carry no claim that one path is more accurate.

**Table 9.** ARC prompts only: outputs that differ from the width-1 reference, differing outputs where both state an option letter, letters that changed, and accuracy against the gold letter on the prompts where both state one.

| path | ARC prompts | outputs differ | differing, both letters | letter changed | reference correct | path correct |
|---|---|---|---|---|---|---|
| 4-token verification | 77 | 60 | 52 | 0 | 22/66 | 22/66 |
| padded to 64 | 77 | 64 | 53 | 2 | 22/65 | 22/65 |
| batch of 4 | 77 | 64 | 56 | 0 | 23/67 | 23/67 |
| gate stale q99 | 50 | 37 | 31 | 0 | 14/44 | 14/44 |
| gate rebuild q99 | 50 | 0 | 0 | 0 | 14/44 | 14/44 |

## 13 Limitations

Everything ran on one GPU (RTX 4090), one driver and one software stack (PyTorch 2.11, transformers 5.20, default SDPA attention). Which widths share a kernel is a property of that stack, so the grouping in the width scan (Section 8) may differ on other hardware or libraries, and so may every divergence rate. Qwen3-0.6B is the main model and Qwen3-1.7B a small replication; larger models may have wider logit gaps and different reduction sizes, and the rates could move either way. The only drafter is prompt lookup; a trained drafter changes which widths occur but not the arguments about cache entries and kernel groups. Generations are 128 tokens with the chat template and thinking switched off, and only greedy decoding was studied, so sampling and long reasoning traces are outside the evidence. Rates describe the 128-token window, and Section 4 shows that they grow with length.

The machine was shared with about eight other jobs and runs were cut to fit. The noise bound was set on 42 calibration prompts (the plan was 120), so the 99th percentile of S rests on about 5253 positions and the 99.9th percentile and the maximum on a handful of them. The bound is an empirical quantity with no guarantee for unseen prompts, and Section 9 tests it on 78 further calibration-set prompts and 60 shifted prompts. The rebuild gate ran on 100 prompts and the other variants on 30 to 100, so their intervals are wide: zero differences in 100 prompts is consistent with a true rate up to about 4%.

Cost is measured mainly as target forward passes per generated token, which assumes decoding is memory-bound. Wall-clock time (Section 7) comes from one run of 12 prompts on a shared GPU and is context only. A pass of width 5 does more arithmetic than a pass of width 1, so passes are a lower bound on cost. The rebuild fallback recomputes the whole generated cache at every fallback, which is the simplest exact implementation and not the cheapest one. An implementation that rewrote each entry once would cost less, and we did not build it. The early-exit read-out is a logit lens on an unmodified model with no trained exit head, and the low-bit weights are simulated round-to-nearest quantisation dequantised to bfloat16, so neither is a deployed kernel.

A divergence is a different token sequence, not a worse answer. We graded only the option letters of ARC outputs, with a regular expression and on a few dozen prompts per path (Section 12), so we cannot say how often a divergence changes correctness on other tasks. The reference is a convention: Section 10 compares paths with a float32 model on 43 prompts and does not show the reference to be more accurate. We did not examine how ties between equal bfloat16 logits are broken, although 39.2% of the pooled first divergences sit at an exact tie.

## 14 Conclusion

Greedy decoding is the same decoding only up to rounding. On Qwen3-0.6B in bfloat16, 4-token verification changed 60.0% [52.0, 67.5] (90/150) of 128-token outputs relative to a width-1 cached reference, padding changed 68.0% [60.2, 74.9] (102/150), and every first divergence sat at a reference margin of 0.375 or less. The noise comes from the shape of the pass and from cache entries that wide passes wrote. A margin gate whose fallback reads those entries did not restore exactness (58.0% [48.2, 67.2] (58/100) still differ), even when an oracle chose the fallback positions (49.0% [39.4, 58.7] (49/100)). Rebuilding the cache with width-1 passes did restore it (0.0% [0.0, 3.7] (0/100) differ) at 7.2 target passes per token, more than the reference spends. Fixed-shape decoding gave agreement between drafted and plain decoding at one width (100 of 100 prompts identical) at 0.75 passes per token, which is one way to keep speculation lossless in floating point. The smallest bound that kept the rebuild gate exact, 0.125 (one step of the bfloat16 grid), cost 3.6 passes per token, and the calibrated bound was about 2.0 times as expensive. On ARC prompts and on shifted MMLU and HellaSwag prompts, which the bound did not see, the stale gate again failed and the rebuild gate left 0 of 60 shifted prompts different. The width-1 path is a convention, and Section 10 shows it is no closer to float32 than the other paths are. A serving stack that needs reproducible greedy output should fix the pass shape, state which shape defines the reference, and treat equality with a different shape as a claim that needs its own check.

## Appendix A  Protocol and commands

Results are cached per path in results/a15b_*.json (shard files .s0, .s1, .s2 are merged by the loaders), and this paper recomputes every number from those files at build time. Three worker processes ran in parallel on the shared GPU, each with two decode threads. Paths: a15b_core.py (verification with gates, fixed-shape decoding, teacher-forced noise sources), drivers a15b_run.py, statistics a15b_data.py, figures a15b_fig.py. The commands are:

```
cd tasks\localllm
# queue 1: teacher-forced noise sources (calibration), then generation on the evaluation prompts
.\a15b_shard.ps1 K 3        (K = 0, 1, 2; tf calib 120 prompts, stopped after 42)
.\a15b_shard2.ps1 K 3 150    (ref, spec4, pad, bs4 on 150 prompts, refb on 60)
# queue 2: gate variants, fixed-shape decoding, width scan, smaller bounds
.\a15b_q2c.ps1 K 3
.\a15b_q2d.ps1 K 3
# queue 3: shifted prompts, float32, cheap paths, second model
.\a15b_q3b.ps1 K 3 60
# queue 4: bound with eps = 0, noise sources on shifted and further calibration prompts, finishing runs
.\a15b_q4.ps1 K 3
# queue 5 and 6: second model, larger samples for two rebuild gates; timing (context only)
.\a15b_q5.ps1
.\a15b_q6.ps1 K 3
python a15b_run.py time 0.6B eval 12 1
```

The text was drafted and edited with LLM assistance; all numbers are recomputed from the stored results by the build scripts and were checked by the authors. The GPU was shared with other users, so wall-clock numbers are context only.

## Appendix B  Noise bounds

The bounds are taken from the teacher-forced calibration run of the 4-token verification path.

**Table 10.** Noise bounds taken from the calibration set (42 prompts, 5253 positions): quantiles and maximum of S for 4-token verification.

| bound | value of eps |
|---|---:|
| q99 | 0.3750 |
| q999 | 0.6250 |
| max | 1.3750 |

## Appendix C  Worked examples

Each row is a prompt that diverged. Tokens are decoded for reading, and positions count generated tokens from zero.

**Table 11.** First divergences in bfloat16: the eight reference tokens before the divergence, the reference token and the path's token, and the two margins.

| path | prompt | position | context | reference | path | ref. margin | path margin |
|---|---:|---:|---|---|---|---:|---:|
| 4-token verification | 0 | 95 | ... Two-thirds of the remainder are to | her | family | 0.250 | 0.000 |
| 4-token verification | 1 | 41 | ... to the **time** when the speed | is | changes | 0.125 | 0.000 |
| 4-token verification | 4 | 72 | ... context of a river ecosystem, the ** | ab | o | 0.000 | 0.125 |
| padded to 64 | 0 | 95 | ... Two-thirds of the remainder are to | her | family | 0.250 | 0.000 |

| path | prompt | position | context | reference | path | ref. margin | path margin |
|---|---|---|---|---|---|---|---|
| padded to 64 | 1 | 41 | ... to the **time** when the speed | is | changes | 0.125 | 0.000 |
| padded to 64 | 3 | 7 | ...We are given: - The price | she | at | 0.000 | 0.375 |
| gate stale q99 | 0 | 24 | ... emails per day Robyn sends ** | 1 | six | 0.125 | 0.000 |
| gate stale q99 | 1 | 41 | ... to the **time** when the speed | is | changes | 0.125 | 0.000 |
| gate stale q99 | 4 | 31 | ...-living factors** that influence the environment | , | and | 0.125 | 0.000 |